\documentclass[lettersize,journal]{IEEEtran}
\usepackage{amsmath,amsfonts}
\usepackage{mathrsfs}
\usepackage{amssymb}
\usepackage{algorithmic}
\usepackage{algorithm}
\usepackage{array}
\usepackage[caption=false,font=normalsize,labelfont=sf,textfont=sf]{subfig}
\usepackage{textcomp}
\usepackage{stfloats}
\usepackage{url}
\usepackage{verbatim}
\usepackage{graphicx}
\usepackage{cite}
\usepackage{makecell}
\usepackage{rotating}
\usepackage{multirow}
\usepackage{longtable}
\usepackage{boldline}
\usepackage{booktabs}
\begin{document}

\title{RLCNet: An end-to-end deep learning framework for simultaneous online calibration of LiDAR, RADAR, and Camera}

\author{Hafeez Husain Cholakkal, Stefano Arrigoni, Francesco Braghin

\thanks{Hafeez Husain Cholakkal, Stefano Arrigoni, and Francesco Braghin are with the Department of Mechanical Engineering, Politecnico di Milano, 20156, Milano, Italy (e-mail: hafeezhusain.cholakkal@polimi.it; stefano.arrigoni@polimi.it; francesco.braghin@polimi.it).}}



\maketitle

\begin{abstract}
Accurate extrinsic calibration of LiDAR, RADAR, and camera sensors is essential for reliable perception in autonomous vehicles. Still, it remains challenging due to factors such as mechanical vibrations and cumulative sensor drift in dynamic environments. This paper presents RLCNet, a novel end-to-end trainable deep learning framework for the simultaneous online calibration of these multimodal sensors. Validated on real-world datasets, RLCNet is designed for practical deployment and demonstrates robust performance under diverse conditions. To support real-time operation, an online calibration framework is introduced that incorporates a weighted moving average and outlier rejection, enabling dynamic adjustment of calibration parameters with reduced prediction noise and improved resilience to drift. An ablation study highlights the significance of architectural choices, while comparisons with existing methods demonstrate the superior accuracy and robustness of the proposed approach.
\end{abstract}

\begin{IEEEkeywords}
Extrinsic calibration, sensor fusion, online calibration, deep learning, lidar, RADAR, camera, message passing network, sensor drift.
\end{IEEEkeywords}

\section{\textbf{Introduction}}

\IEEEPARstart{A}{utonomous} vehicles are poised to revolutionize transportation by improving road safety, reducing traffic congestion, and increasing mobility convenience \cite{hussain2019autonomous}. To perceive and interact with their environment accurately, these vehicles rely on a combination of complementary sensors, including LiDAR, RADAR, and cameras. Each sensor offers unique advantages: cameras capture rich visual detail, LiDAR provides precise 3D spatial measurements, and RADAR performs robustly under adverse weather conditions \cite{yeong2021sensor}. Sensor fusion leverages the strengths of these modalities to ensure redundancy and resilience, allowing the vehicle to maintain accurate perception in diverse and dynamic environments \cite{wang2019multi}.

A critical component of sensor fusion is extrinsic calibration, which involves the determination of the relative positions and orientations of sensors in a common coordinate frame. However, maintaining precise calibration over time is a persistent challenge. Factors such as mechanical vibrations, temperature changes, and minor collisions can lead to sensor drift, where even small misalignments in sensor orientation or position can result in substantial perception errors, potentially compromising vehicle safety.

This paper addresses the problem of simultaneous online extrinsic calibration of LiDAR, RADAR, and camera sensors in autonomous vehicles. Unlike traditional pairwise calibration methods, which are often time-consuming and prone to inconsistency, we propose a unified framework that calibrates all sensor pairs jointly. This approach not only improves efficiency but also ensures loop closure consistency among all sensor pairs, enhancing overall system robustness. We assume access to temporally aligned and intrinsically calibrated sensor data, as intrinsic calibration is typically managed by built-in software, and temporal synchronization resolves misalignment across data streams. Building on these assumptions allows us to isolate the extrinsic calibration problem, ensuring that our framework focuses solely on estimating the relative poses between sensors without the confounding effects of intrinsic errors or temporal misalignment.

The proposed method, RLCNet, is an end-to-end trainable deep learning architecture that predicts transformation errors in the extrinsic parameters, enabling real-time calibration adjustments based on incoming sensor data. RLCNet is validated using publicly available datasets collected from real-world driving environments, ensuring practical applicability and demonstrating its effectiveness in handling real-world calibration challenges.
The contributions of this work are threefold:

\begin{itemize}
    \item \textbf{End-to-End Trainable Deep Learning Approach for Simultaneous Online Calibration of LiDAR, RADAR, and Camera Sensors:}
    
    We introduce RLCNet, a novel deep network that jointly calibrates LiDAR, RADAR, and camera sensors by estimating extrinsic transformation errors. Unlike existing approaches, RLCNet enables online, targetless calibration with minimal supervision and is well-suited for real-time applications.
    
    \item \textbf{Message-Passing Optimization for Loop Closure Consistency:}

    RLCNet incorporates a message passing network (MPN) module to enhance the consistency of extrinsic parameters across all sensor pairs. The network is trained end-to-end with a custom loss function that penalizes accuracy degradation, promoting global calibration coherence.
    
    \item \textbf{Online Calibration Framework with Temporal Filtering and Outlier Rejection:}

    To address temporal dependencies in calibration drift, we design a lightweight online monitoring system that applies a weighted moving average to smooth the predictions while remaining responsive to gradual drift. An outlier detection mechanism filters anomalous predictions, enhancing reliability in dynamic conditions. We also outline a drift detection and recalibration strategy to maintain long-term calibration accuracy.
\end{itemize}

The remainder of this paper is organized as follows. Section~\ref{sec:literature} reviews related work in the field of sensor calibration for autonomous vehicles. Section \ref{sec:methodology} describes the architecture of the proposed RLCNet framework and outlines the training methodology. In Section \ref{sec:experiments}, we present a comprehensive evaluation of RLCNet, including performance comparisons with state-of-the-art approaches. The design and analysis of the online calibration framework are detailed in Section \ref{sec:online}. Finally, Section \ref{sec:conclusions} summarizes the key findings and contributions of this research.

\section{\textbf{Related Works}}\label{sec:literature}

Extrinsic calibration methods for multi-sensor systems can generally be classified into two main categories: target-based and targetless approaches. Target-based methods utilize known calibration objects or markers placed within the sensors’ shared field of view to establish spatial correspondences. In contrast, targetless methods rely on natural environmental features or scene geometry, offering greater flexibility and practicality in unstructured, real-world conditions.

\subsection{\textbf{Target-Based Calibration}}

Target-based calibration methods are grounded in the use of physical markers or objects that are observable by all sensors involved. These known references simplify the task of finding accurate correspondences across modalities, enabling precise estimation of relative sensor poses. Such methods are especially effective because they provide strong geometric constraints and reduce ambiguities during calibration.

In the early development of LiDAR-camera calibration techniques, checkerboard patterns were commonly employed. Some approaches require manual annotation of corresponding keypoints across sensor views \cite{lyu2019interactive}, while others automate the process using line and plane correspondences \cite{zhou2018automatic}. Planar boards with circular holes have also proven effective for LiDAR-camera calibration \cite{guindel2017automatic}. More sophisticated targets combining visual and geometric features, such as checkerboards integrated with holes, have been proposed to improve robustness across modalities \cite{beltran2022automatic}. Additionally, spherical targets have been introduced to offer greater flexibility in sensor placement and orientation \cite{kummerle2018automatic}.

Calibrating RADAR with other sensors poses unique challenges due to RADAR's lower resolution and its sensitivity to reflective surface properties. A key design consideration for RADAR targets is the RADAR Cross Section (RCS), which determines the strength of the RADAR return signal. Various target types have been proposed, including RADAR-detectable ArUco markers \cite{song2017novel} and styrofoam boards embedded with steel spheres \cite{wirth2024automatic}. Despite these developments, the trihedral corner reflector remains the most widely adopted RADAR calibration target due to its consistently high RCS and ease of detection \cite{kim2019extrinsic, cheng20233d, pervsic2017extrinsic}.

Efforts have also been made to develop targets suitable for the simultaneous calibration of LiDAR, RADAR, and camera systems. However, the challenge lies in designing patterns that are reliably detectable across all sensing modalities. Some approaches have employed styrofoam boards with multiple circular holes alongside trihedral corner reflectors \cite{domhof2021joint}, while others have used multiple rigidly connected targets combining distinct visual features with corner reflectors to facilitate joint calibration \cite{agrawal2023static}.

Although target-based methods offer high accuracy due to their controlled setup and well-defined environments, they are inherently limited in their ability to address sensor drift during operation. For autonomous systems to function reliably under dynamic real-world conditions, it is essential that they can detect and recover from calibration errors online and without human intervention.

\subsection{\textbf{Targetless Calibration}}
Targetless calibration methods derive correspondences directly from features present in the natural environment, eliminating the need for predefined calibration targets. These approaches leverage inherent geometric structures such as edges, corners, planes, and other salient scene features to estimate the relative poses between sensors. By relying on naturally occurring elements rather than artificial markers, targetless methods offer greater flexibility and adaptability, making them particularly well-suited for real-world, unstructured, and dynamic environments.

One category within the domain of targetless calibration is based on information theory, where statistical similarity measures are computed from sensor data to optimize extrinsic parameters. For instance, LiDAR surface reflectivity and camera grayscale intensity can be aligned using mutual information \cite{pandey2015automatic, borer2024chaos}. To improve robustness against global intensity variations and color space shifts, normalized information distance has been employed \cite{pascoe2015direct, koide2023general}.

\begin{figure*}[t]
    \centering
    \includegraphics[width=\textwidth]{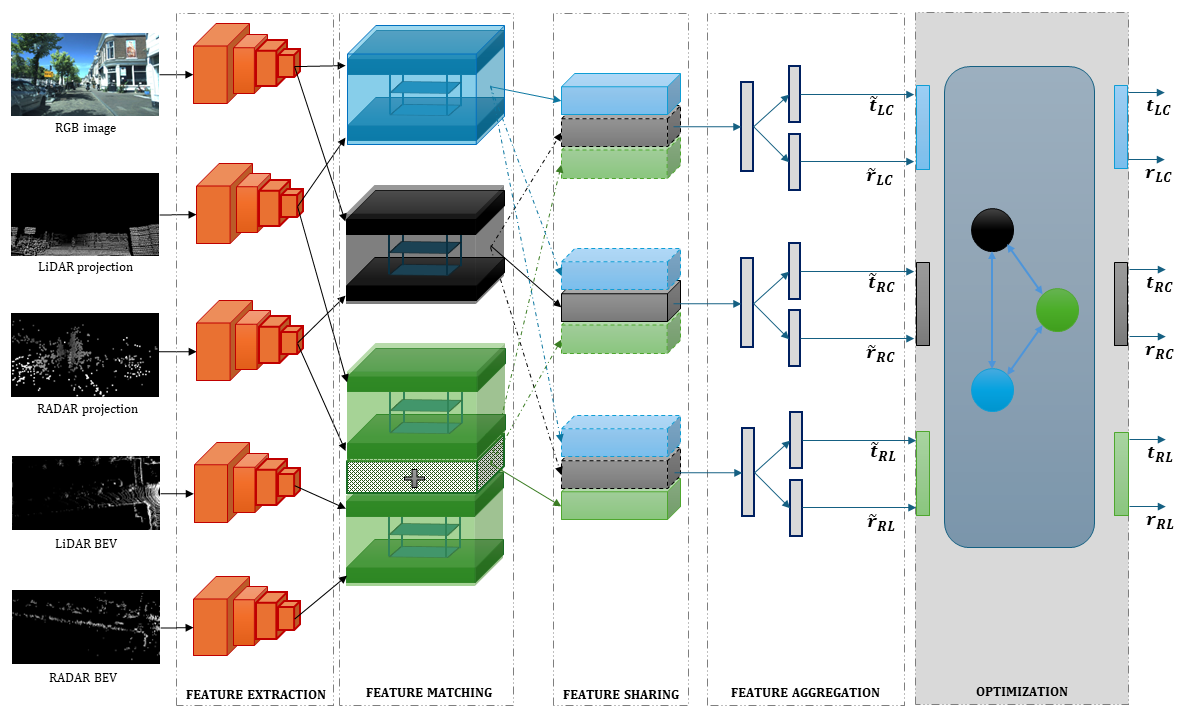}
    \caption{Network architecture overview of RLCNet}
    \label{fig:rlcnet}
\end{figure*}

A more intuitive class of targetless methods is feature-based calibration, which relies on identifying stable geometric or semantic features, such as edges, lines, and objects, for cross-modal matching. Common techniques include maximizing edge alignment between projected LiDAR data and camera images \cite{levinson2013automatic, zhang2021line}, as well as using line segmentation for 2D RADAR-LiDAR calibration \cite{rotter2022calibration}. Semantic features like vehicles have also been used for online LiDAR-camera calibration \cite{zhu2020online}.

Another approach involves ego-motion-based calibration, where sensor trajectories are estimated independently and then aligned to recover extrinsic parameters. This technique has been applied to both LiDAR-camera \cite{taylor2016motion, horn2021online} and RADAR-camera systems \cite{wise2023spatiotemporal}.

Recently, deep learning-based methods have gained traction for targetless calibration. These models learn feature representations directly from data, eliminating the need for handcrafted features. Examples include RegNet \cite{schneider2017regnet}, LCCNet \cite{lv2021lccnet}, and CalibRCNN \cite{shi2020calibrcnn}, the latter integrating convolutional features with LSTM modules to exploit temporal consistency. Similar learning-based approaches have also been proposed for RADAR-camera calibration \cite{cheng2023online}.

A few works have explored simultaneous targetless calibration of LiDAR, RADAR, and camera. Per{\v{s}}i{\'c} et al. \cite{pervsic2021online} proposed a method based on tracking semantic features (e.g., moving vehicles) across modalities, focusing on correcting rotational miscalibration. More recently, Hayoun et al. \cite{hayoun2024physics} introduced two unified calibration approaches: an optimization-based method and a self-supervised learning (SSL) framework. Both utilize semantic features to establish pairwise constraints and a global loop closure condition. In the optimization-based approach, these constraints are used to define a loss function optimized over multiple synchronized frames. In the SSL approach, sensor-specific encoders extract features that are fed into a deep network trained to regress the relative transformations between sensors. While these approaches demonstrate strong potential, their overall effectiveness remains closely tied to the accuracy of the diverse methodologies employed for semantic feature extraction. In contrast, our method is trained end-to-end and operates as a standalone system, requiring only a synchronized feed from the sensors to perform online calibration.

\section{\textbf{Methodology}}\label{sec:methodology}

This section outlines the proposed calibration framework and the associated training process. As a preliminary step, we assume that all sensors in the suite are pre-calibrated for their intrinsic parameters, as this is a fundamental prerequisite for the effective operation of RLCNet. Intrinsic calibration is typically performed by the sensor manufacturer. In the case of cameras, periodic recalibration may be necessary to maintain accuracy, which can be efficiently achieved using target-based methods. Accordingly, we consider the camera intrinsic matrix $\mathbf{K}$ to be known and accurate.

\subsection{\textbf{Input Processing}}\label{sec:input_processing}
The input to RLCNet comprises three components: the RGB image from the camera, projection images of the LiDAR and RADAR point clouds in the camera frame, and Bird's Eye View (BEV) representations of both LiDAR and RADAR data.

To generate these inputs, synchronized frames from all sensors are first acquired. The LiDAR and RADAR point clouds are then transformed into the camera coordinate system using the initial extrinsic estimates, denoted as $\mathbf{T}_{CL}^{\text{init}}$ (LiDAR-to-camera) and $\mathbf{T}_{CR}^{\text{init}}$ (RADAR-to-camera). Using the camera’s intrinsic matrix $\mathbf{K}$, projection images are created for the transformed point clouds. Each pixel in these images encodes the inverse depth of the corresponding 3D point.

To generate the bird’s eye view (BEV), it is first necessary to define the lateral and longitudinal ranges of the point clouds selected from the LiDAR and RADAR sensors. This choice is guided by the sensors’ effective detection ranges, resolution, and their shared field of view. For the dataset used in our experiments, we found a lateral range of $[-15 \text{m}, 15 \text{m}]$ and a longitudinal range of $[0 \text{m}, 60 \text{m}]$ to be optimal. The point clouds are then filtered to retain only points within this region of interest, after which the filtered data is rotated to align the $XY$ plane with the image plane $(u,v)$.

Each 3D point $\mathbf{p}_i = [x_i, y_i, z_i]$ is projected to 2D pixel coordinates $(u_i, v_i)$ using the following mapping:
\begin{align}
u_i &= \textit{scale}_x \cdot x_i + \textit{offset}_x, \\
v_i &= \textit{scale}_y \cdot y_i + \textit{offset}_y,
\end{align}
where $\textit{scale}_x$ and $\textit{scale}_y$ are scaling factors determined by the target input resolution of the images. In our implementation, we set $\textit{scale}_x = \textit{scale}_y = 10$, which corresponds to a spatial resolution of $0.1\,\text{m} \times 0.1\,\text{m}$ per pixel. The offsets are computed based on the defined spatial ranges:
\begin{align}
\textit{offset}_x &= -\textit{range}_{\text{lat}}[0] / \textit{scale}_x, \\
\textit{offset}_y &= -\textit{range}_{\text{long}}[1] / \textit{scale}_y.
\end{align}

This process yields a BEV representation in which each pixel encodes the height of its corresponding 3D point. Given the selected region of interest and spatial resolution, the resulting images have a resolution of $600 \times 300$ pixels.

\subsection{\textbf{Network Architecture}}\label{sec:architecture}

RLCNet consists of five key modules: \textit{feature extraction}, \textit{feature matching}, \textit{feature sharing}, \textit{feature aggregation}, and \textit{optimization}. An overview of the network architecture is illustrated in Figure~\ref{fig:rlcnet}.

\subsubsection{\textbf{Feature Extraction Module}}

The feature extraction module in RLCNet comprises five parallel convolutional branches. For the RGB image branch, we utilize a pre-trained ResNet-18 architecture \cite{resnet}, omitting its fully connected layers. To retain low-level semantic features, the weights of the first two convolutional layers are kept frozen during training.

The depth and BEV image branches share a common architecture based on a modified ResNet-18. These branches are adapted to accept single-channel inputs by setting the number of input channels to one. Furthermore, standard ReLU activations are replaced with Leaky ReLU to improve gradient flow and feature representation in low-intensity regions. All input images are resized to a resolution of $256 \times 512$ before being fed into the network.

\subsubsection{\textbf{Feature Matching Module}}
The feature matching module computes pairwise matching costs to quantify the similarity between feature representations extracted from different sensor modalities. Specifically, we employ three parallel feature matching layers for all three pairwise combinations of RGB and depth images, with each layer computing a correlation cost volume between the corresponding feature maps. A fourth matching layer computes the correlation cost between features extracted from LiDAR and RADAR BEV images.

The correlation cost between a pixel \( p_1 \) in the feature map from sensor 1 (\( \mathbf{x}_{s1} \)) and a pixel \( p_2 \) in the feature map from sensor 2 (\( \mathbf{x}_{s2} \)) is defined as:
\begin{equation}
    cc(p_1, p_2) = \frac{1}{N} \left( \mathbf{c}(\mathbf{x}_{s1}(p_1)) \right)^T \mathbf{c}(\mathbf{x}_{s2}(p_2)),
\end{equation}
where \( \mathbf{c}(\mathbf{x}_{si}(p_j)) \) is the flattened feature vector at pixel \( p_j \) in the feature map from sensor \( i \), and \( N \) is the dimensionality of the feature vector.

For each \( p_1 \), the correlation cost is computed with respect to a local neighborhood of candidate pixels \( p_2 \) in the other feature map, constrained by a maximum displacement \( d \):
\begin{equation}
|p_1 - p_2|_{\infty} \leq d.
\end{equation}
The resulting correlation values are stored in a \emph{3D cost volume} (\( \mathbf{cv} \)) with dimensions \( (2d+1)^2 \times H \times W \), where \( H \) and \( W \) are the spatial dimensions of the feature maps.

To exploit the complementary nature of projection and BEV representations, the cost volumes from these two modalities are concatenated in the feature space. As a result, the cost volume constructed between LiDAR and RADAR features has dimensions \( 2(2d+1)^2 \times H \times W \).

Due to downsampling from convolutional layers, the feature maps are reduced to a resolution of \( 8 \times 16 \). To maintain computational efficiency, we empirically choose \( d = 3 \) as the optimal displacement threshold. Finally, all cost volumes are projected into 1D arrays of uniform size using multilayer perceptrons (MLPs), ensuring consistent dimensionality before passing them to the feature sharing module.

\subsubsection{\textbf{Feature Sharing Module}}

The feature sharing module is designed to strengthen interdependencies among pairwise calibration parameters by facilitating information exchange across sensor pairs. It takes the three feature branches generated by the feature matching layers, fuses them, and restructures them into three updated branches, which are then passed to the feature aggregation stage.

Inspired by the approach in \cite{dahal2023soft}, we investigate two feature sharing strategies: (a) direct fusion, where features are combined without any weighting, and (b) soft mask fusion, where a learned mask modulates the contribution of each feature during fusion.

\begin{enumerate}
\item[a)] \textbf{Direct Feature Fusion:}

A simple yet effective method for feature fusion involves concatenating the feature vectors from each sensor pair and passing the result through a series of three fully connected layers for feature aggregation. This direct fusion strategy can be mathematically formulated as:
\begin{equation}
    \mathbf{g}^{\text{direct}} = \left[ f\left( \mathbf{cv}_{LC} \right) ; f\left( \mathbf{cv}_{RC} \right) ; f\left( \mathbf{cv}_{RL} \right) \right],
\end{equation}
where \( f\left( \mathbf{cv}_{ij} \right) \) represents the feature vector obtained from the correlation layer between sensor $i$ and $j$, and \([ \cdot \hspace{2pt}; \hspace{2pt}\cdot \hspace{2pt}; \hspace{2pt}\cdot ]\) denotes the concatenation operation.

\begin{figure}[!t]
\centering
\subfloat[Direct fusion]{\includegraphics[width=0.3\linewidth]{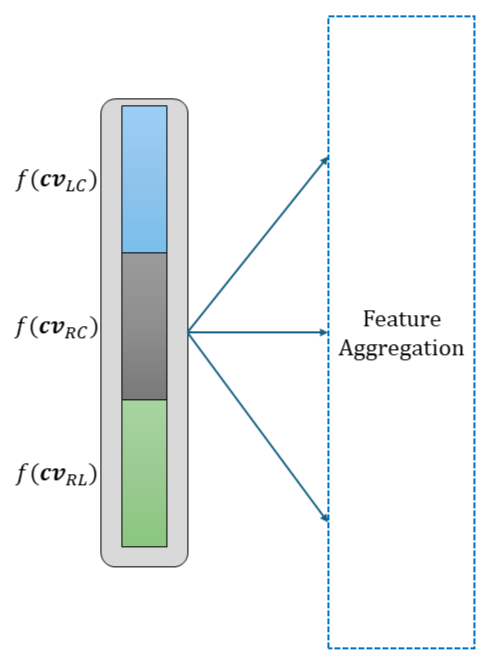}%
\label{fig:direct_fusion}}
\hfil
\subfloat[Soft mask fusion]{\includegraphics[width=0.5\linewidth]{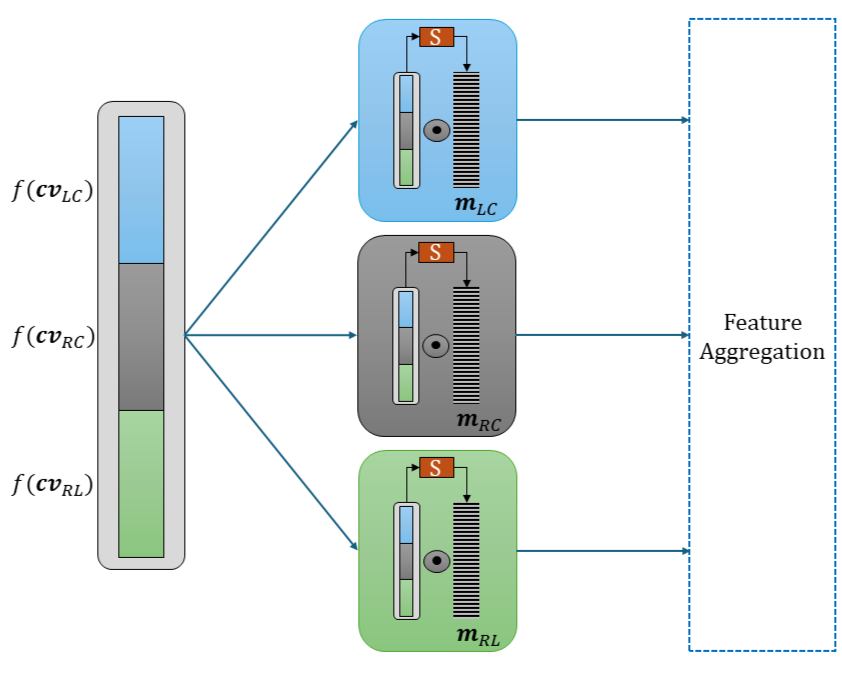}%
\label{fig:soft_mask}}
\caption{An illustration of the two feature sharing schemes}
\label{fig:feature_sharing}
\end{figure}

\item[b)] \textbf{Feature Sharing through Soft Mask:}
The proposed method adopts a soft fusion strategy that explicitly and deterministically models feature sharing across sensor modalities. This is achieved by reweighting each feature based on the full set of feature channels output by the feature matching module. This design enables the feature-sharing process to be jointly trained with other modules in a deterministic and differentiable manner.

To support multi-sensor prediction, the feature-sharing module generates three output branches corresponding to the sensor pairs: LiDAR–Camera, RADAR–Camera, and LiDAR–RADAR. For this purpose, we introduce three continuous masks, denoted as $\mathbf{m}_{LC}$, $\mathbf{m}_{RC}$, and $\mathbf{m}_{RL}$. These masks facilitate feature selection from the concatenated feature vector by emphasizing features relevant to each output branch.

The masks are parametrized deterministically using neural networks and are defined as follows:
\begin{equation}
    \mathbf{m}_{ij} = \mathrm{Sigmoid}_{ij} \left( \mathscr{N} \left[ \mathbf{g}^{\text{direct}} \right] \right), 
\end{equation}
where $\mathscr{N}\left[ \mathbf{g}^{\text{direct}} \right]$ is a multi-layer perceptron (MLP) applied to the concatenated feature vector, and $\mathrm{Sigmoid}_{ij}$ maps the output to the range $[0, 1]$, producing a soft weighting mask for sensor pair $ij$.

The soft masks have the same dimensionality as the concatenated feature vector and are applied element-wise to generate the respective output branches. This can be mathematically represented as:
\begin{equation}
    \mathbf{g}^{\text{soft}}_{ij} = \mathbf{m}_{ij} \odot \mathbf{g}^{\text{direct}}
\end{equation}
An illustration of the feature-sharing schemes is presented in Figure~\ref{fig:feature_sharing}.
\end{enumerate}

\subsubsection{\textbf{Feature Aggregation Module}}

The feature aggregation module of RLCNet comprises three parallel branches of fully connected layers, each dedicated to estimating the calibration parameters for a specific sensor pair. Each branch begins with a 512-neuron layer, which splits into two sub-branches with 256 neurons each. One sub-branch predicts a translation vector $\Tilde{\mathbf{t}} \in \mathbb{R}^{3}$, while the other outputs a rotation quaternion $\Tilde{\mathbf{r}} \in \mathbb{R}^{4}$. These intermediate predictions are then passed to the optimization module for refinement and final calibration.

\subsubsection{\textbf{Optimization Module}}

The intermediate transformations predicted by the feature aggregation module for the LiDAR-camera, RADAR-camera, and LiDAR-RADAR sensor pairs are denoted as $\Tilde{\mathbf{T}}_{LC}$, $\Tilde{\mathbf{T}}_{RC}$, and $\Tilde{\mathbf{T}}_{RL}$, respectively. However, these initial predictions may not satisfy the loop closure constraint:

\begin{equation*}
\Tilde{\mathbf{T}}_{RL} \neq \Tilde{\mathbf{T}}_{RC} \cdot (\Tilde{\mathbf{T}}_{LC})^{-1}.
\end{equation*}

To enforce consistency, we introduce a \emph{Message Passing Network (MPN)} within the optimization module. The MPN refines the predicted transformations by iteratively propagating information across sensor pairs to minimize loop closure error. This module is fully differentiable and trained end-to-end in conjunction with the rest of the network.

\textbf{Graph Structure:} The proposed Message-Passing Network (MPN) consists of three nodes, each representing the transformation between a pair of sensors. Node features are initialized using intermediate transformation estimates. Due to the inherent interdependence among these transformations, enforced by the loop closure constraint, the graph forms a fully connected triangular structure, with edges connecting every pair of nodes to facilitate information exchange.

\textbf{Message Passing:} In a message-passing framework, each node exchanges information with its neighboring nodes via the edges of a graph. The message received by a node can be interpreted as the expected transformation that satisfies the loop closure constraint with respect to its neighbors. Specifically, we define the following messages:
\begin{align}
    m_{LC} &= \left( \Tilde{\mathbf{T}}_{RL} \right)^{-1} \cdot \Tilde{\mathbf{T}}_{RC} \\[5pt]
    m_{RC} &= \Tilde{\mathbf{T}}_{RL} \cdot \Tilde{\mathbf{T}}_{LC} \\[5pt]
    m_{RL} &= \Tilde{\mathbf{T}}_{RC} \cdot \left( \Tilde{\mathbf{T}}_{LC} \right)^{-1}
\end{align}

\textbf{Node Update:} At each iteration of the message-passing process, the node features are updated based on the messages received from neighboring nodes. The update rule is defined as:

\begin{align}
    \Tilde{\mathbf{T}}_{LC}^{(t)} &= \alpha_t \cdot \Tilde{\mathbf{T}}_{LC}^{(t-1)} + (1 - \alpha_t) \cdot m_{LC}^{(t-1)} \\
    \Tilde{\mathbf{T}}_{RC}^{(t)} &= \alpha_t \cdot \Tilde{\mathbf{T}}_{RC}^{(t-1)} + (1 - \alpha_t) \cdot m_{RC}^{(t-1)} \\
    \Tilde{\mathbf{T}}_{RL}^{(t)} &= \alpha_t \cdot \Tilde{\mathbf{T}}_{RL}^{(t-1)} + (1 - \alpha_t) \cdot m_{RL}^{(t-1)}
\end{align}
where $\alpha_t$ is a set of learnable hyperparameters of the MPN that controls the trade-off between the current transformation estimate and the incoming message at iteration $t$. 

Given the relatively small number of nodes, four message-passing iterations are sufficient for RLCNet to achieve loop closure consistency and refine the transformation estimates. After the final iteration, the network outputs the calibrated transformations, represented as a translation vector $\mathbf{t}^{\text{pred}}$ and a rotation quaternion $\mathbf{r}^{\text{pred}}$.

\subsection{\textbf{Loss Function}}

RLCNet estimates the deviation between the initial extrinsic transformation and the ground truth transformation for each sensor pair. The network is fully end-to-end trainable and optimized via supervised learning. Its training objective is defined as a weighted combination of multiple loss components: the \textit{pose loss} $L_p$, the \textit{point cloud distance loss} $L_c$, the \textit{loop closure loss} $L_l$, and \textit{accuracy penalty} $L_a$.

\begin{figure*}[t]
    \centering
    \includegraphics[width=\textwidth]{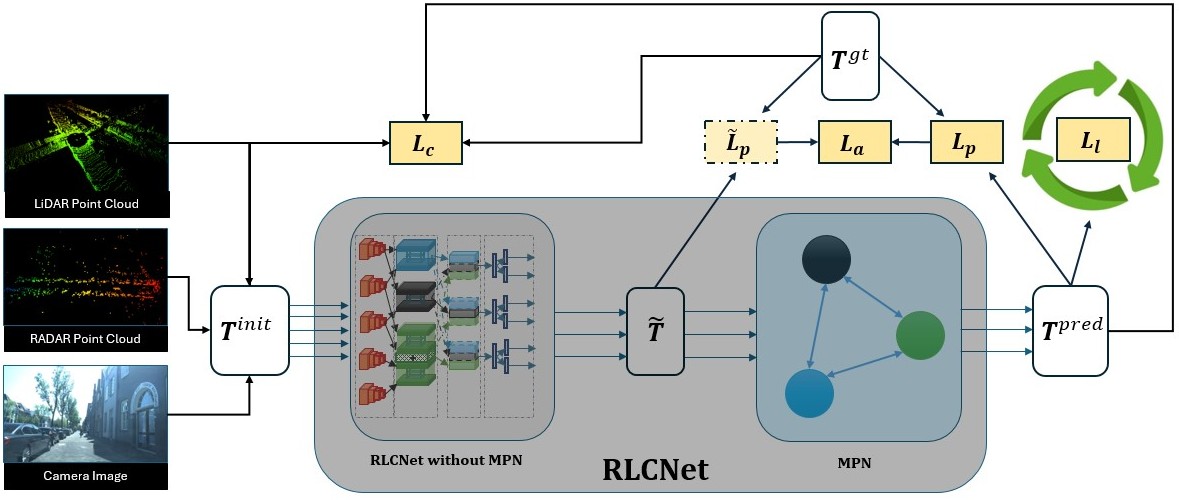}
    \caption{Workflow of training RLCNet}
    \label{fig:loss_function}
\end{figure*}

\subsubsection{\textbf{Pose Loss}}

The pose loss measures the discrepancy between predicted and ground-truth extrinsic transformations across all sensor pairs. It combines both rotational and translational components, each weighted appropriately:

\begin{equation}\label{eq:pose_loss}
L_p = \lambda_r \cdot L_r + \lambda_t \cdot L_t
\end{equation}

The translation loss measures the difference between the predicted translation vector $\mathbf{t}^{\text{pred}}$ and the ground truth translation vector $\mathbf{t}^{\text{gt}}$, computed using the Smooth L1 loss across the $x$, $y$, and $z$ components:

\begin{equation}
L_t = \sum_{i=1}^3 \text{Smooth L1}(\mathbf{t}_i^{\text{gt}}, \mathbf{t}_i^{\text{pred}})
\end{equation}

As for rotation, quaternions encode directional information on a unit sphere, making Euclidean distance an unsuitable metric. Instead, the rotation loss is defined using the angular distance between the predicted quaternion $\mathbf{r}^{\text{pred}}$ and the ground truth quaternion $\mathbf{r}^{\text{gt}}$:

\begin{equation}
L_r = \sum_{i=1}^3 D_q(\mathbf{r}_i^{\text{gt}}, \mathbf{r}_i^{\text{pred}})
\end{equation}

where $D_q$ represents the angular distance, defined as the minimal angle of rotation along the shortest arc connecting the two quaternions on the hypersphere. $D_q$ between two unit quaternions $\mathbf{r}^{\text{gt}}$ and $\mathbf{r}^{\text{pred}}$ can be expressed explicitly as:

\begin{equation}
D_q(\mathbf{r}^{\text{gt}}, \mathbf{r}^{\text{pred}}) = 2 \cdot \arccos\left( \left| \langle \mathbf{r}^{\text{gt}}, \mathbf{r}^{\text{pred}} \rangle \right| \right),
\end{equation}
where $\langle \mathbf{r}^{\text{gt}}, \mathbf{r}^{\text{pred}} \rangle$ is the dot product between the quaternions.

\subsubsection{\textbf{Point Cloud Distance Loss}}
The point cloud distance loss measures the average Euclidean distance between each point in the original point cloud and its transformed counterpart, using the estimated extrinsic calibration parameters. This loss encourages the predicted transformation to align the point cloud with the ground truth frame as closely as possible.

Given a LiDAR point cloud $\mathbf{P}_l = \{ \mathbf{p}_1, \mathbf{p}_2, \dots, \mathbf{p}_{N_l} \} \in \mathbb{R}^3$, the average point cloud distance loss is formulated as:
\begin{equation}
    L_c^{\text{LiDAR}} = \frac{1}{N_l} \sum_{i=1}^{N_l} \left\lVert \mathbf{T}_{LC}^{\text{gt}} \cdot \left( \mathbf{T}_{LC}^{\text{pred}} \right)^{-1} \cdot \mathbf{T}_{CL}^{\text{init}} \cdot \hat{\mathbf{p}}_i - \hat{\mathbf{p}}_i \right\rVert_2,
\end{equation}
where:
\begin{itemize}
    \item $\mathbf{T}_{LC}^{\text{gt}}$ is the ground-truth transformation matrix from the camera to the LiDAR,
    \item $\mathbf{T}_{LC}^{\text{pred}}$ is the predicted transformation matrix from the camera to the LiDAR,
    \item $\mathbf{T}_{CL}^{\text{init}}$ is the initial transformation matrix from the LiDAR to the camera,
    \item $N_l$ is the number of points in the LiDAR point cloud,
    \item $\hat{\mathbf{p}}_i$ denotes the homogeneous coordinates of the $i$-th point $\mathbf{p}_i$,
    \item $\lVert \cdot \rVert_2$ denotes the Euclidean ($L^2$) norm.
\end{itemize}

A similar formulation is applied to the RADAR point cloud to compute $L_c^{\text{RADAR}}$. The total point cloud distance loss is given by:
\begin{equation}
    L_c = L_c^{\text{LiDAR}} + L_c^{\text{RADAR}}.
\end{equation}

\subsubsection{\textbf{Loop Closure Loss}}

Loop closure loss encourages consistency among the predicted extrinsic transformations by enforcing the closed-loop constraint among sensor pairs. The loop closure estimates of the extrinsic transformations are computed as:

\begin{align}
    \Hat{\mathbf{T}}_{LR}^{\text{loop}} &= \Hat{\mathbf{T}}_{CL}^{-1} \cdot \Hat{\mathbf{T}}_{CR}, \label{eq:loop_estimations1} \\[5pt]
    \Hat{\mathbf{T}}_{CL}^{\text{loop}} &= \Hat{\mathbf{T}}_{CR} \cdot \Hat{\mathbf{T}}_{LR}^{-1}, \\[5pt]
    \Hat{\mathbf{T}}_{CR}^{\text{loop}} &= \Hat{\mathbf{T}}_{CL} \cdot \Hat{\mathbf{T}}_{LR}. \label{eq:loop_estimations3}
\end{align}

Loop closure loss \( L_l \) is defined as the average pose loss between the directly predicted transformations and their corresponding loop closure estimates:

\begin{equation}
    L_l = \frac{1}{3} \sum_{i=1}^3 L_p'\left(\Hat{\mathbf{T}}_{i}^{\text{loop}}, \Hat{\mathbf{T}}_{i}\right),
\end{equation}

where \( L_p' \) denotes the pose loss function (as defined in Equation~\ref{eq:pose_loss}) evaluated between the predicted transformations \( \Hat{\mathbf{T}}_i \) and their loop-based counterparts \( \Hat{\mathbf{T}}_i^{\text{loop}} \).

\subsubsection{\textbf{Accuracy Penalty}}

To ensure that the optimization process carried out by the MPN does not degrade the accuracy of the predictions, we incorporate an \textit{accuracy penalty} term into the total loss function. This penalty is applied only when the final refined prediction is less accurate than the intermediate prediction. Specifically, we compute the pose loss for the intermediate predictions, denoted as $\tilde{\mathbf{t}}$ and $\tilde{\mathbf{r}}$, with respect to the ground truth:
\begin{align}
    \tilde{L}_r &= \sum_{i=1}^{3} D_q(\mathbf{r}_i^{\text{gt}}, \tilde{\mathbf{r}}_i) \\[5pt]
    \tilde{L}_t &= \sum_{i=1}^{3} \text{Smooth L1}(\mathbf{t}_i^{\text{gt}}, \tilde{\mathbf{t}}_i) \\[5pt]
    \tilde{L}_p &= \lambda_r \cdot \tilde{L}_r + \lambda_t \cdot \tilde{L}_t
\end{align}

The accuracy penalty is then defined as the positive difference between the pose loss of the final predictions and that of the intermediate predictions:

\begin{equation}
    L_a = \text{ReLU}(L_p - \tilde{L}_p)
\end{equation}

Here, the ReLU function ensures that a penalty is only incurred if the refined predictions result in a higher error (i.e., lower accuracy) compared to the intermediate predictions.

Figure~\ref{fig:loss_function} illustrates the various components that contribute to the overall loss. The complete loss function is given by:

\begin{equation}
    L = (1 - (\lambda_c + \lambda_l)) \cdot L_p + \lambda_c \cdot L_c + \lambda_l \cdot L_l + \gamma \cdot L_a
\end{equation}

where $\lambda_c$ and $\lambda_l$ are weighting factors for the point cloud loss ($L_c$) and the loop closure loss ($L_l$), respectively. The coefficient $\gamma$ determines the contribution of the accuracy penalty ($L_a$) to the total loss.

\subsection{\textbf{Iterative Refinement}}

Inspired by recent advances in deep learning-based extrinsic calibration methods~\cite{schneider2017regnet, lv2021lccnet}, RLCNet adopts an iterative refinement strategy to improve calibration accuracy progressively. Specifically, a sequence of RLCNet models is trained, where each subsequent network handles a smaller range of miscalibration. The transformation predicted by each network is used to correct the input data before it is passed to the next stage, enabling increasingly precise predictions at each iteration.

Let the initial transformation predicted by the first network be denoted as $\mathbf{T}^{1}$. Given the original LiDAR and RADAR point clouds in homogeneous coordinates, $\Hat{\mathbf{P}}_l$ and $\Hat{\mathbf{P}}_r$, the transformed point clouds after applying the estimated corrections are given by:

\begin{align}
    \Hat{\mathbf{P}}_l^1 &= \left(\mathbf{T}_{LC}^{1}\right)^{-1} \cdot \mathbf{T}_{CL}^{\text{init}} \cdot \Hat{\mathbf{P}}_l,\\[5pt]
    \Hat{\mathbf{P}}_r^1 &= \left(\mathbf{T}_{RC}^{1}\right)^{-1} \cdot \mathbf{T}_{CR}^{\text{init}} \cdot \Hat{\mathbf{P}}_r.
\end{align}

Following the input processing steps described in Section~\ref{sec:input_processing}, new depth projections and BEV images are generated from the updated point clouds. These, together with the original RGB image, serve as inputs to the next network in the cascade, which predicts a refined transformation $\mathbf{T}^{2}$. Repeating this process iteratively across $n$ networks yields the final camera-to-LiDAR and camera-to-RADAR extrinsic transformations as:
\begin{align}
    \Hat{\mathbf{T}}_{CL} &= \left( \mathbf{T}_{LC}^{1} \cdot \mathbf{T}_{LC}^{2} \cdots \mathbf{T}_{LC}^{n} \right)^{-1} \cdot \mathbf{T}_{CL}^{\text{init}}, \\[5pt]
    \Hat{\mathbf{T}}_{CR} &= \left( \mathbf{T}_{RC}^{1} \cdot \mathbf{T}_{RC}^{2} \cdots \mathbf{T}_{RC}^{n} \right)^{-1} \cdot \mathbf{T}_{CR}^{\text{init}}.
\end{align}

However, this formulation cannot be directly applied to compute the final LiDAR-to-RADAR extrinsic transformation unless the loop closure constraint is strictly satisfied at each iteration. Instead, the refined LiDAR-to-RADAR transformation is computed as:

\begin{equation}
    \Hat{\mathbf{T}}_{LR} = \left(\mathbf{T}_{RL}^{n}\right)^{-1} \cdot \left(\Hat{\mathbf{T}}_{CL}^{n-1}\right)^{-1} \cdot \Hat{\mathbf{T}}_{CR}^{n-1},
\end{equation}

where $\Hat{\mathbf{T}}_{CL}^{n-1}$ and $\Hat{\mathbf{T}}_{CR}^{n-1}$ are the refined extrinsic transformations obtained up to iteration $(n-1)$.

The initial miscalibration ranges are set to $[-10^{\circ}, +10^{\circ}]$ for each rotational degree of freedom and $[-50\,\text{cm}, +50\,\text{cm}]$ for each translational degree of freedom for both the LiDAR–camera and RADAR–camera sensor pairs. For subsequent networks, the miscalibration ranges are reduced based on the maximum mean absolute error (MAE) of the predictions from the previous stage. In total, five networks are trained with decreasing miscalibration ranges defined as:
\begin{align*}
    [-r, r], \hspace{2pt} r &\in \{10^{\circ}, 6^{\circ}, 4^{\circ}, 2^{\circ}, 1^{\circ} \}, \\
    [-t, t], \hspace{2pt} t &\in \{50\,\text{cm}, 30\,\text{cm}, 20\,\text{cm}, 10\,\text{cm}, 5\,\text{cm} \},
\end{align*}
where $r$ and $t$ denote the maximum range of rotational and translational perturbations, respectively.

\subsection{\textbf{Training the Network}}

Given the increasing prevalence of 3D RADARs, we selected the View of Delft (VoD) dataset \cite{vod} for training our network. This dataset provides synchronized frames of high-definition LiDAR and RADAR point clouds, as well as front-facing RGB images. The sensor setup consists of a \textit{ZF FR-Gen 21 RADAR}, an \textit{IDS RGB stereo camera}, and a \textit{Velodyne HDL-64S3 LiDAR}. All sensors were jointly calibrated using the target-based approach described in \cite{domhof2021joint}, offering precise extrinsic parameters for each frame in the dataset. The data is divided into a training set containing 6435 frames and a test set with 2247 frames.

\begin{figure}[!t]
\centering
\includegraphics[width=0.8\linewidth]{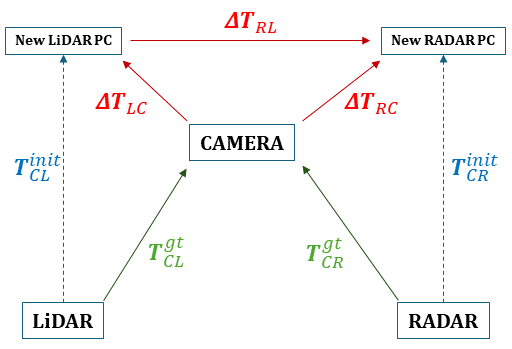}
\caption{An illustration of generated random transformations}
\label{fig:deltat}
\end{figure}

\subsubsection{\textbf{Data Preparation}}

In order to train the networks across different miscalibration ranges, it is essential to have control over the miscalibration of individual frames. This control is achieved by first transforming the LiDAR and RADAR point clouds into the camera coordinate system, using the ground truth transformations from the dataset. A random transformation $\mathbf{\Delta T}$ is then generated within the specified miscalibration range for both LiDAR and RADAR data. These transformations are added to the ground truth to obtain the initial transformations for the network as follows:

\begin{align}
    \mathbf{T}_{CL}^{\text{init}} &= \mathbf{\Delta T}_{LC} \cdot \mathbf{T}_{CL}^{\text{gt}},\\[5pt]
    \mathbf{T}_{CR}^{\text{init}} &= \mathbf{\Delta T}_{RC} \cdot \mathbf{T}_{CR}^{\text{gt}}.
\end{align}

\begin{figure*}[!t]
\centering
\subfloat[LiDAR input]{\includegraphics[width=0.32\textwidth]{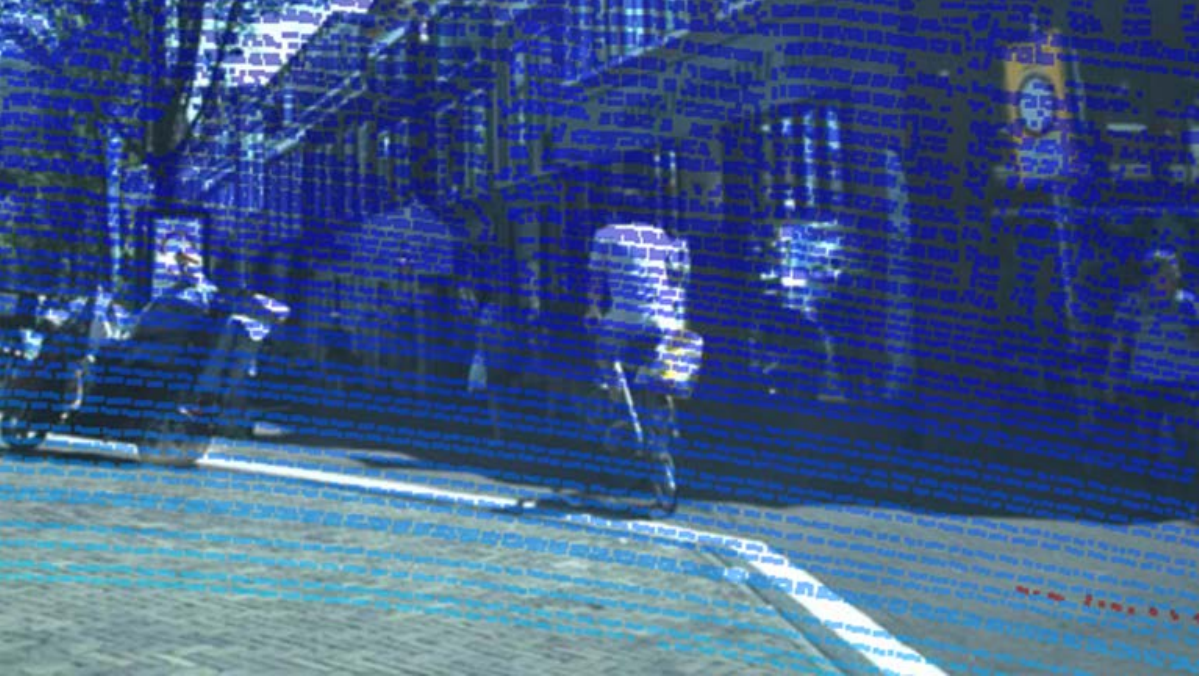}%
\label{fig:scene2 lid in}}
\hfil
\subfloat[LiDAR ground truth]{\includegraphics[width=0.32\textwidth]{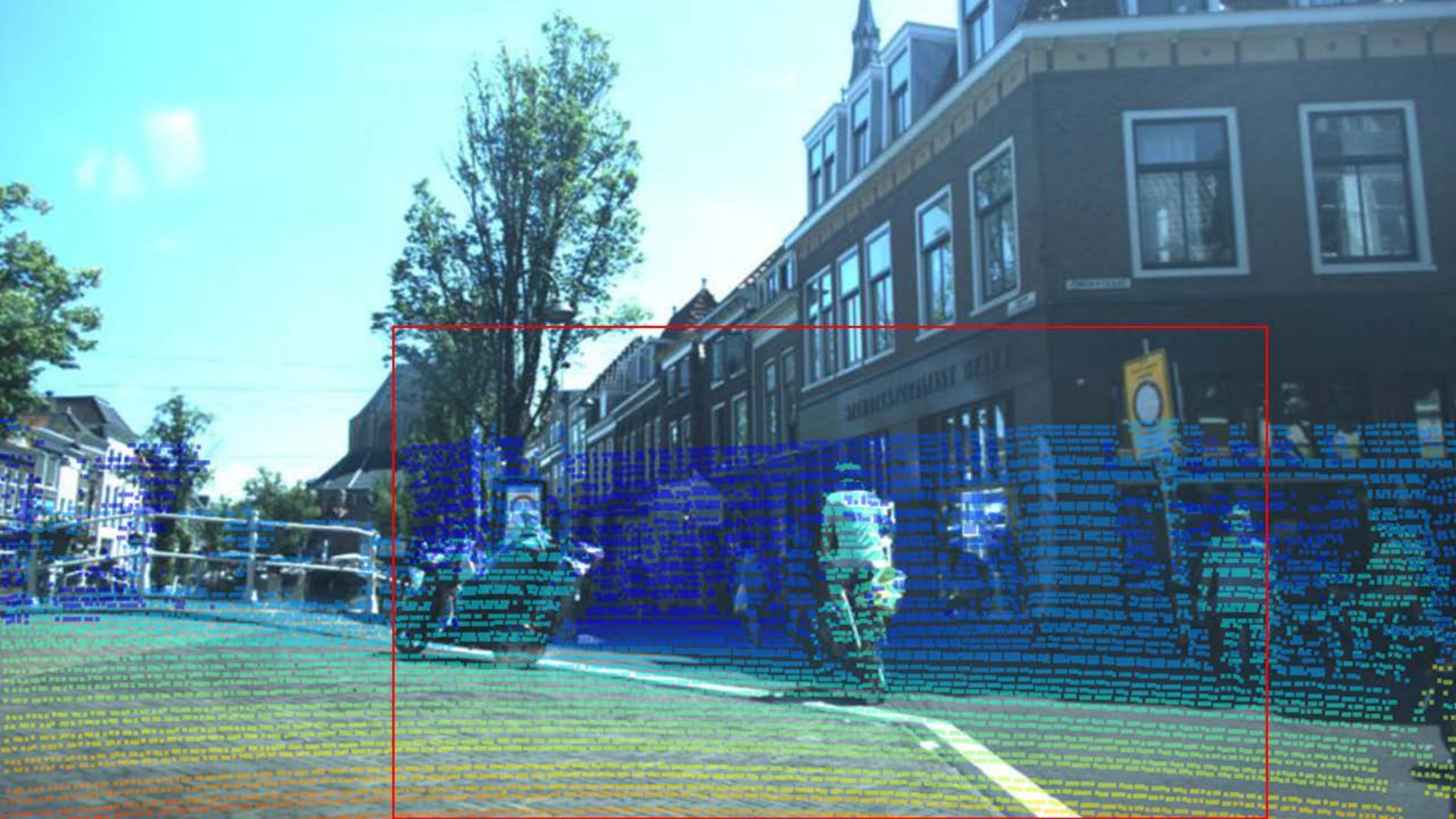}%
\label{fig:scene2 lid gt}}
\hfil
\subfloat[LiDAR prediction]{\includegraphics[width=0.32\textwidth]{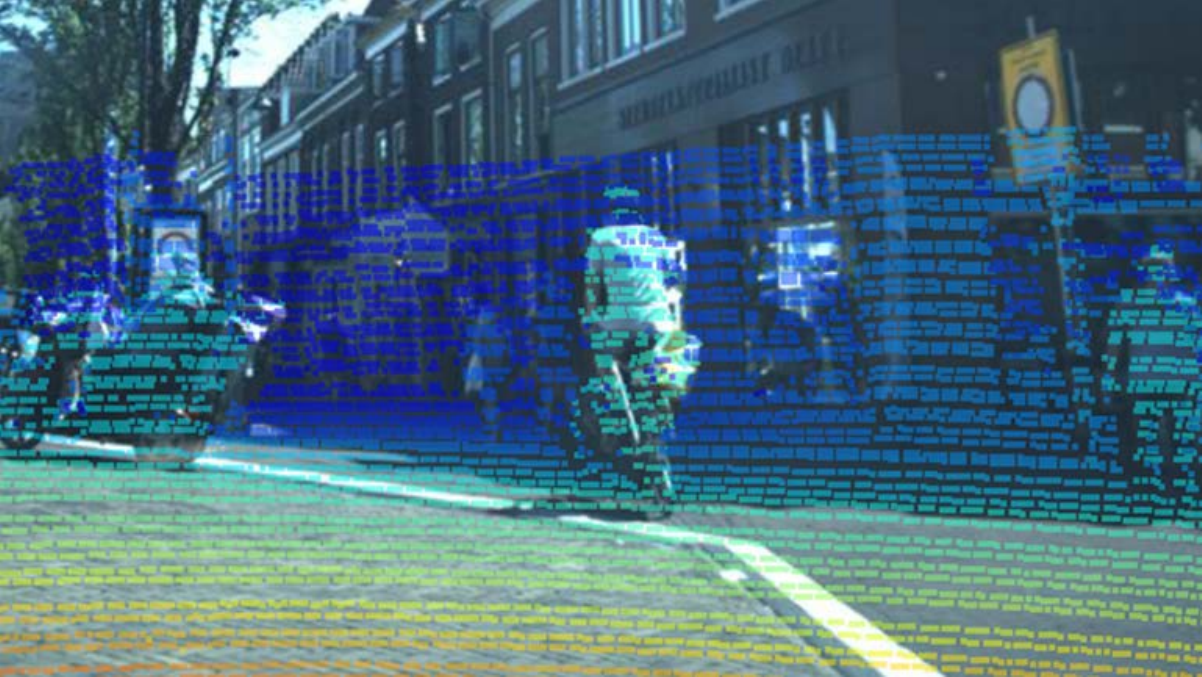}%
\label{fig:scene2 lid pred}}

\subfloat[RADAR input]{\includegraphics[width=0.32\textwidth]{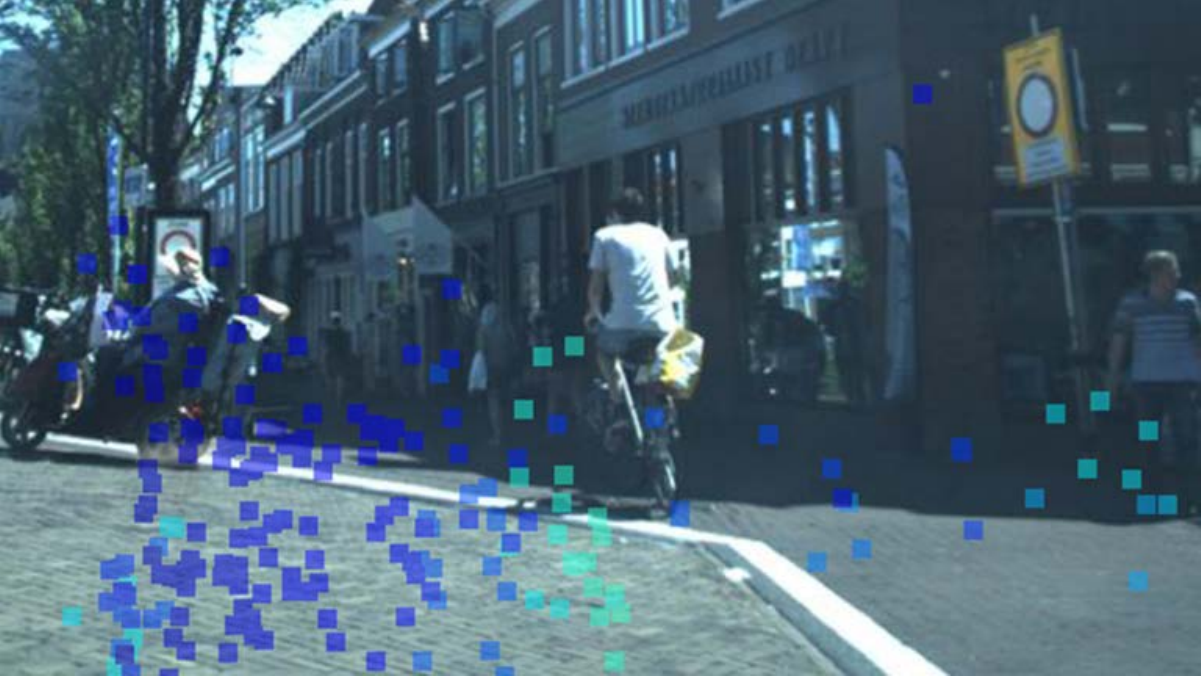}%
\label{fig:scene2 rad in}}
\hfil
\subfloat[RADAR ground truth]{\includegraphics[width=0.32\textwidth]{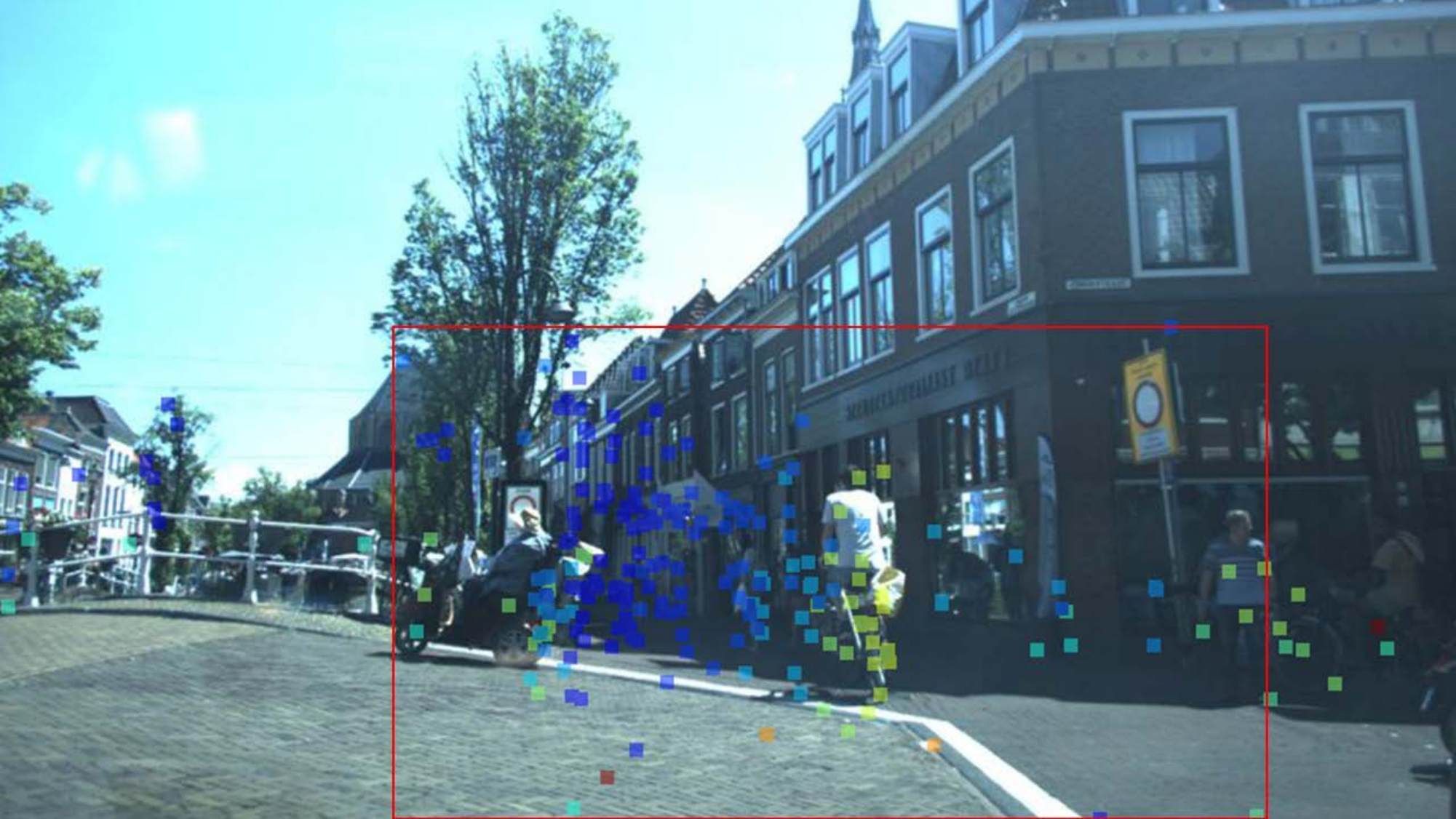}%
\label{fig:scene2 rad gt}}
\hfil
\subfloat[RADAR prediction]{\includegraphics[width=0.32\textwidth]{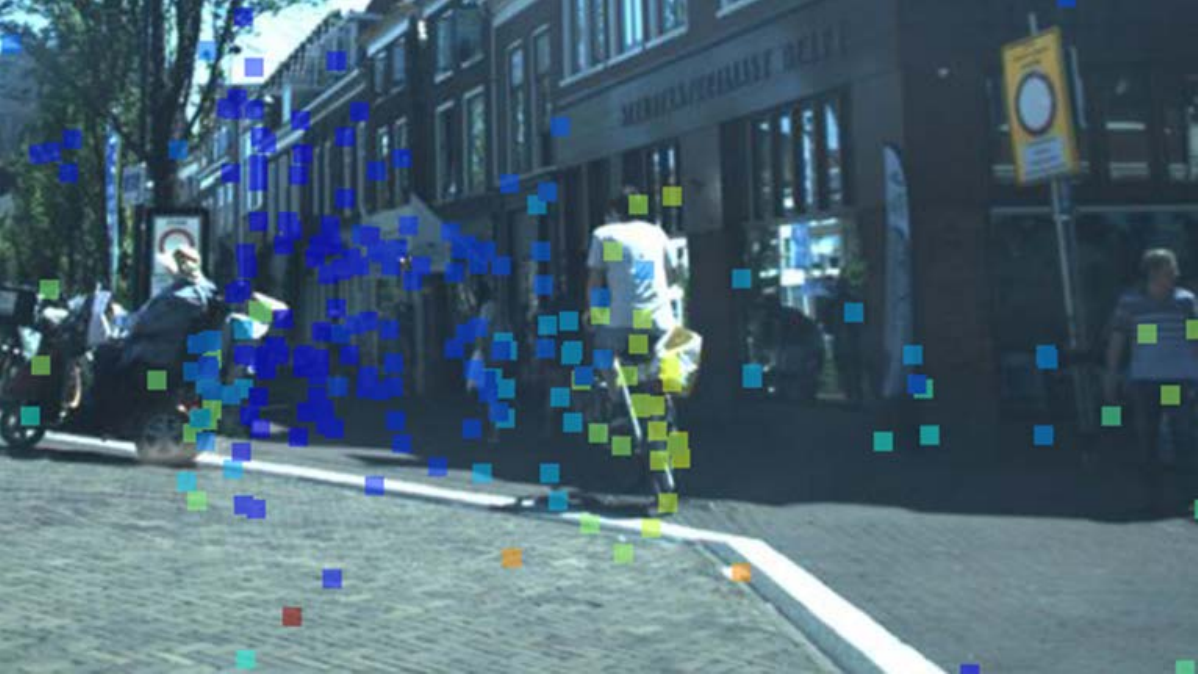}%
\label{fig:scene2 rad pred}}

\subfloat[BEV input]{\includegraphics[width=0.32\textwidth]{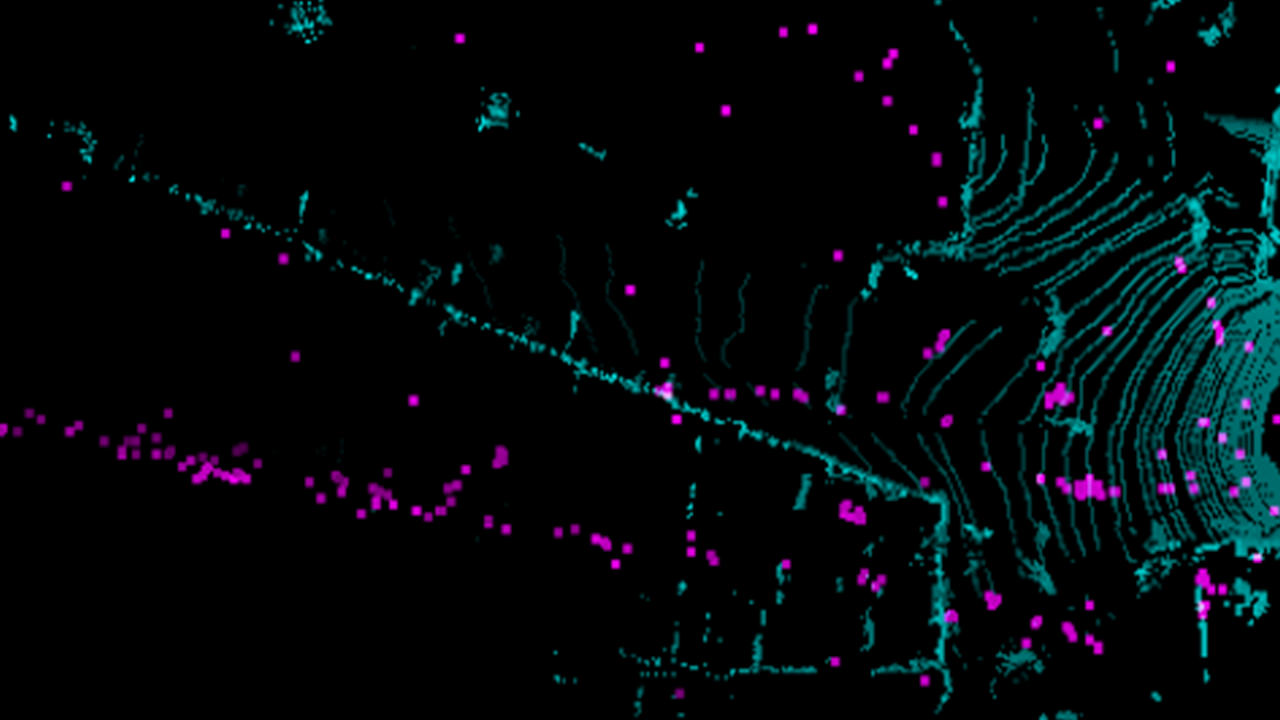}%
\label{fig:scene2 bev in}}
\hfil
\subfloat[BEV ground truth]{\includegraphics[width=0.32\textwidth]{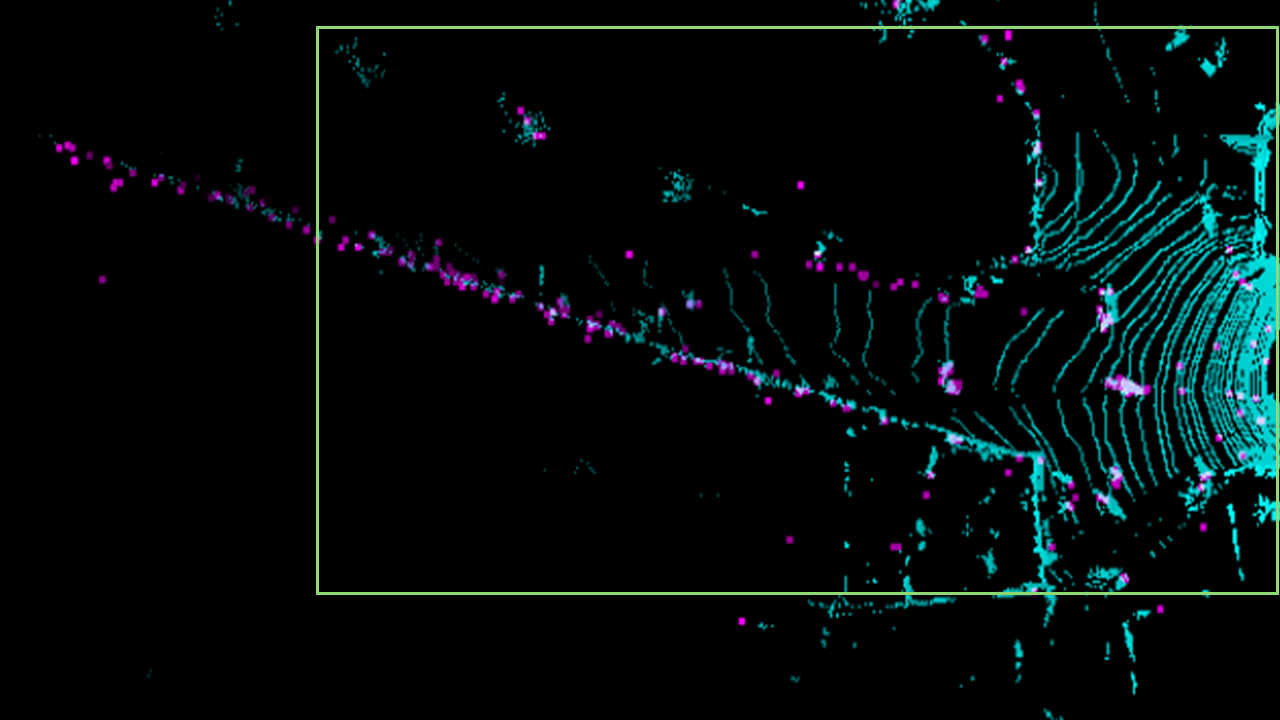}%
\label{fig:scene2 bev gt}}
\hfil
\subfloat[BEV prediction]{\includegraphics[width=0.32\textwidth]{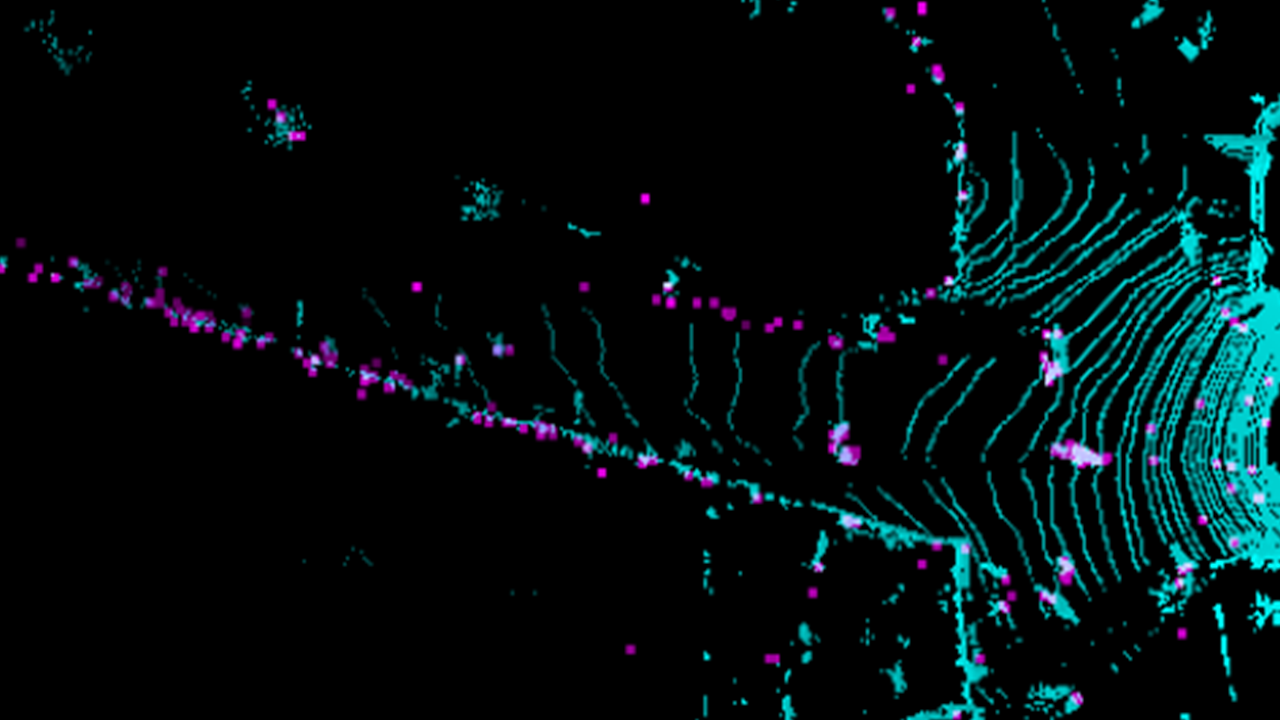}%
\label{fig:scene2 bev pred}}
\caption{Calibration results of RLCNet after iterative refinement (Scene 1)}
\label{fig:rlcnet_results1}
\end{figure*}
\begin{figure*}[!t]
\centering
\subfloat[LiDAR input]{\includegraphics[width=0.32\textwidth]{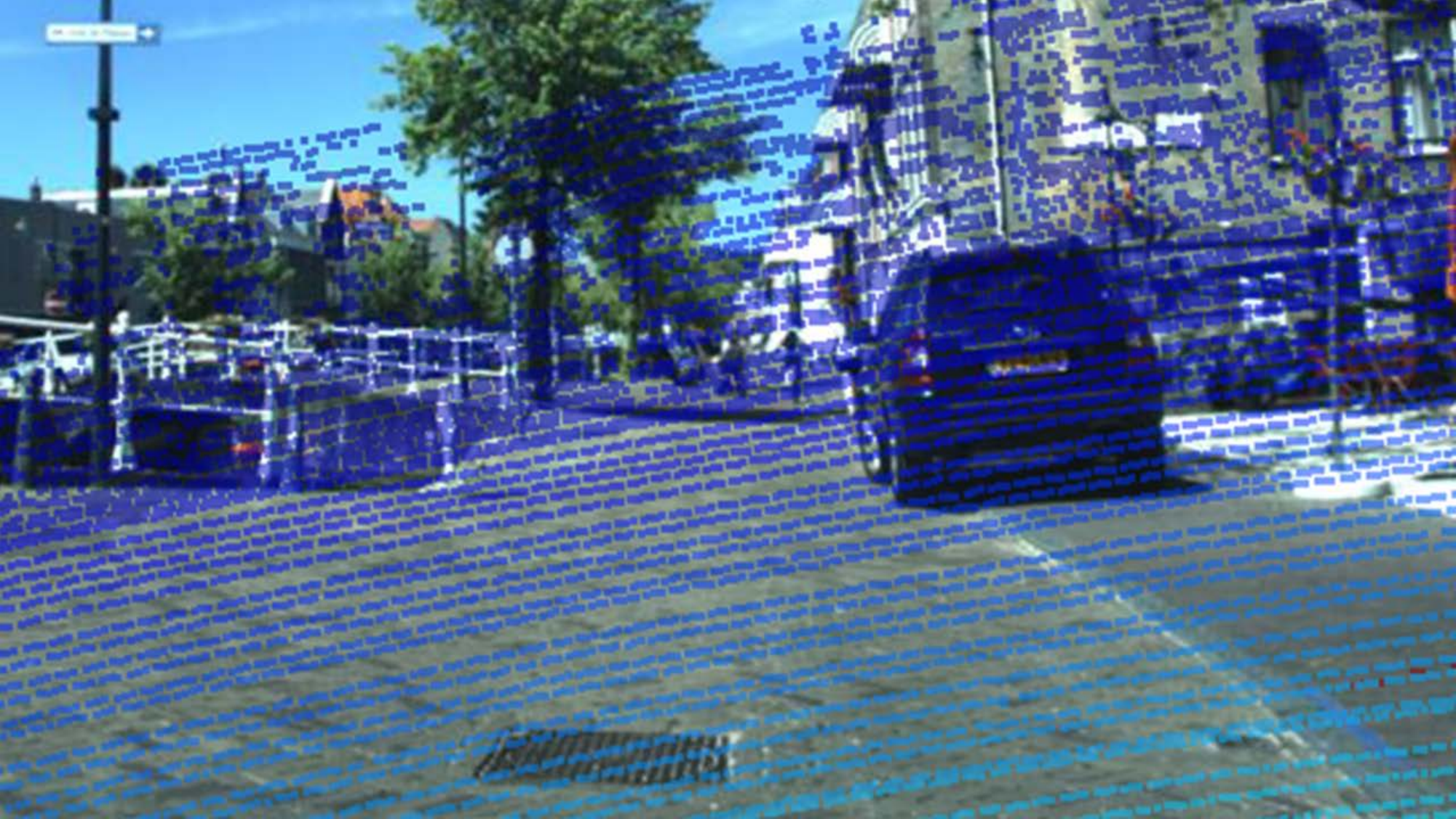}%
\label{fig:scene3 lid in}}
\hfil
\subfloat[LiDAR ground truth]{\includegraphics[width=0.32\textwidth]{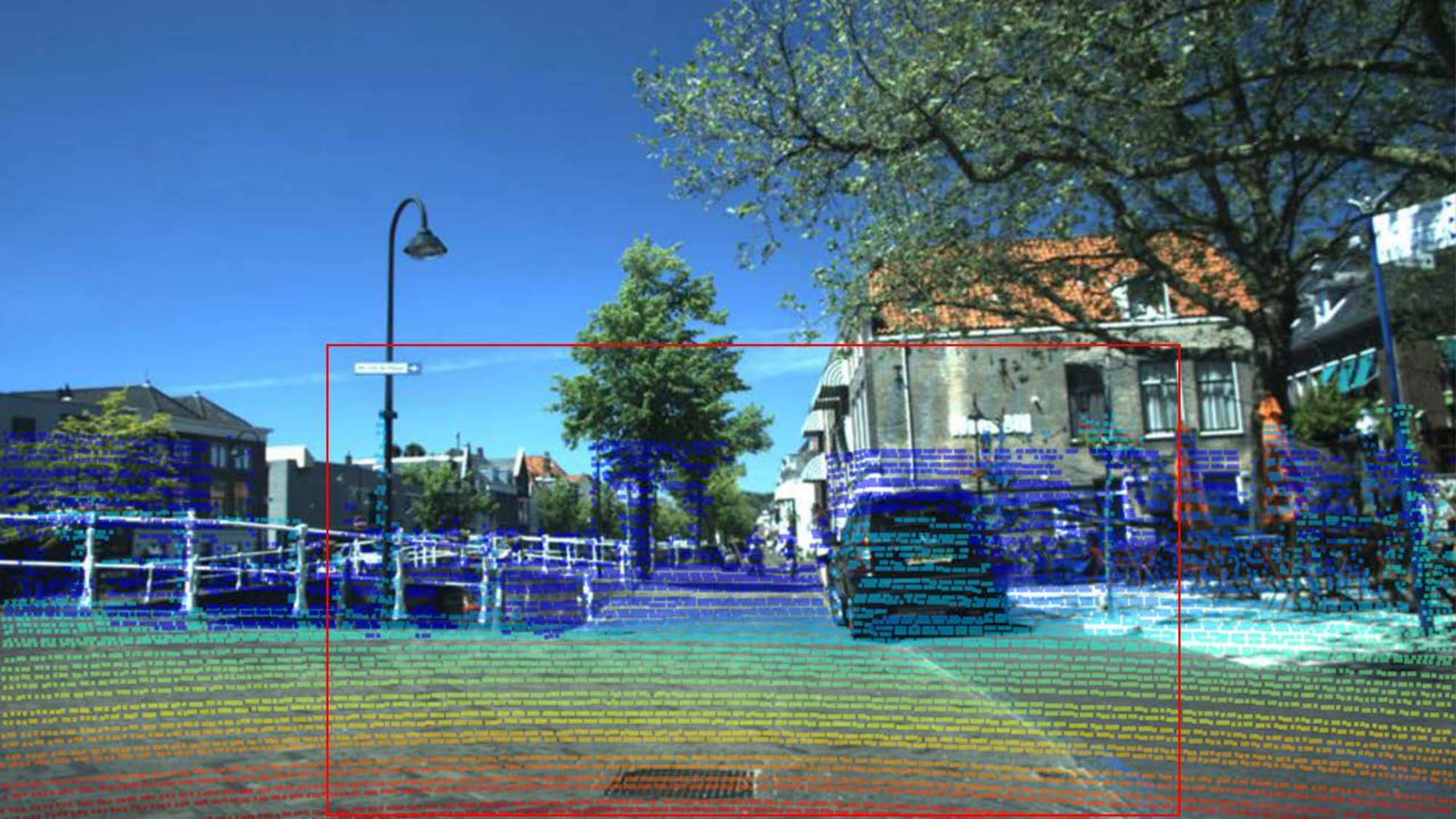}%
\label{fig:scene3 lid gt}}
\hfil
\subfloat[LiDAR prediction]{\includegraphics[width=0.32\textwidth]{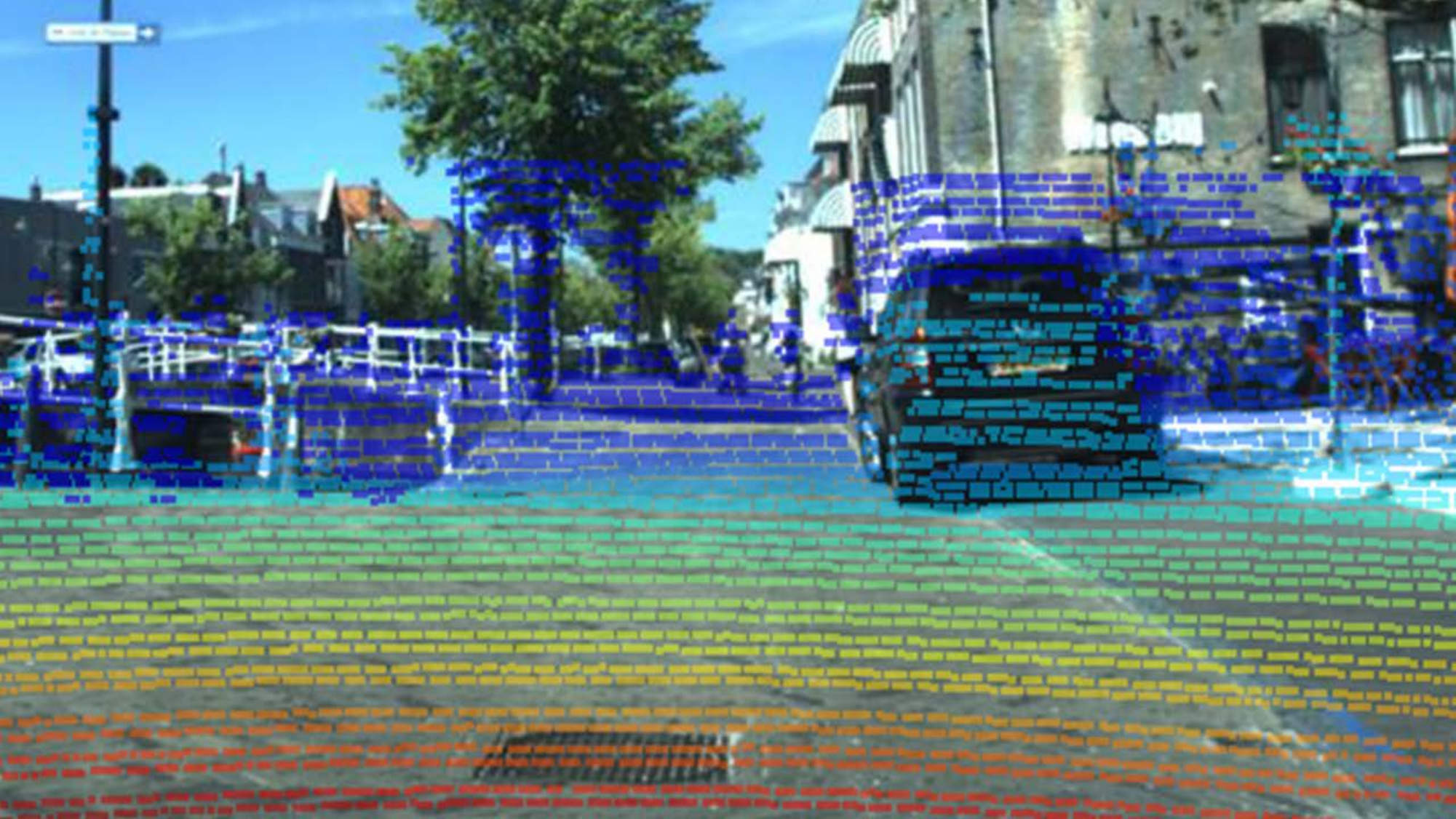}%
\label{fig:scene3 lid pred}}

\subfloat[RADAR input]{\includegraphics[width=0.32\textwidth]{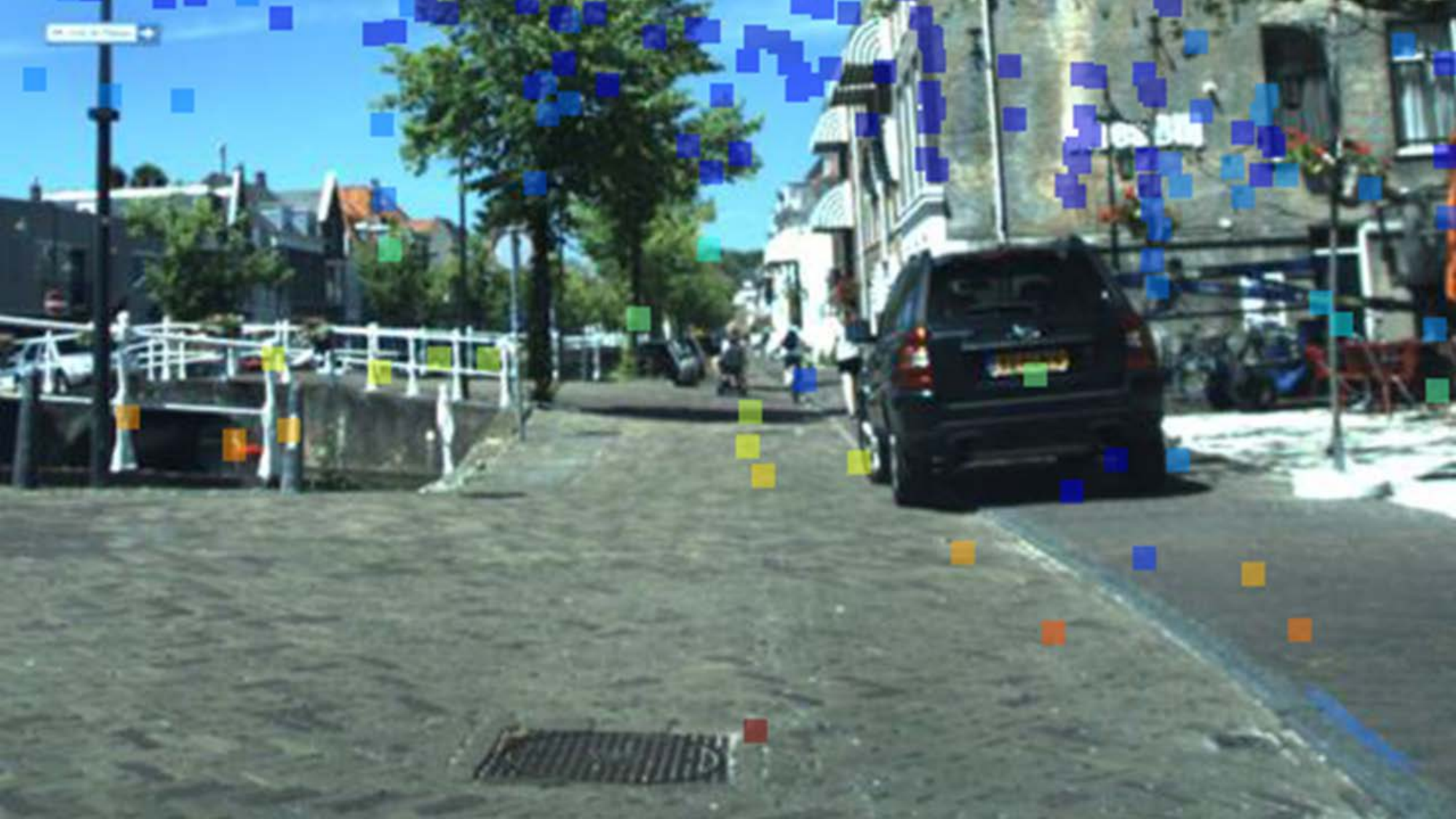}%
\label{fig:scene3 rad in}}
\hfil
\subfloat[RADAR ground truth]{\includegraphics[width=0.32\textwidth]{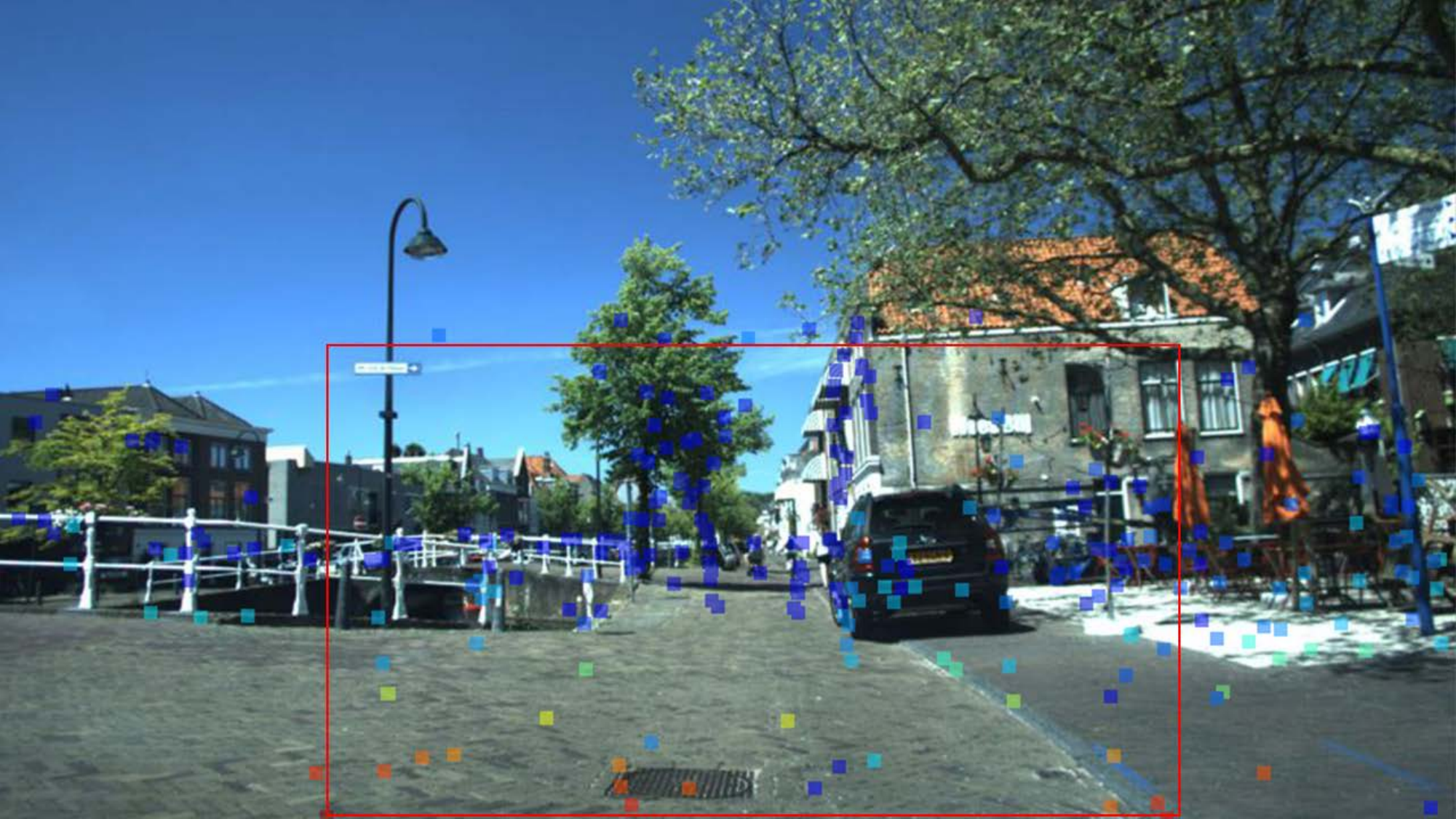}%
\label{fig:scene3 rad gt}}
\hfil
\subfloat[RADAR prediction]{\includegraphics[width=0.32\textwidth]{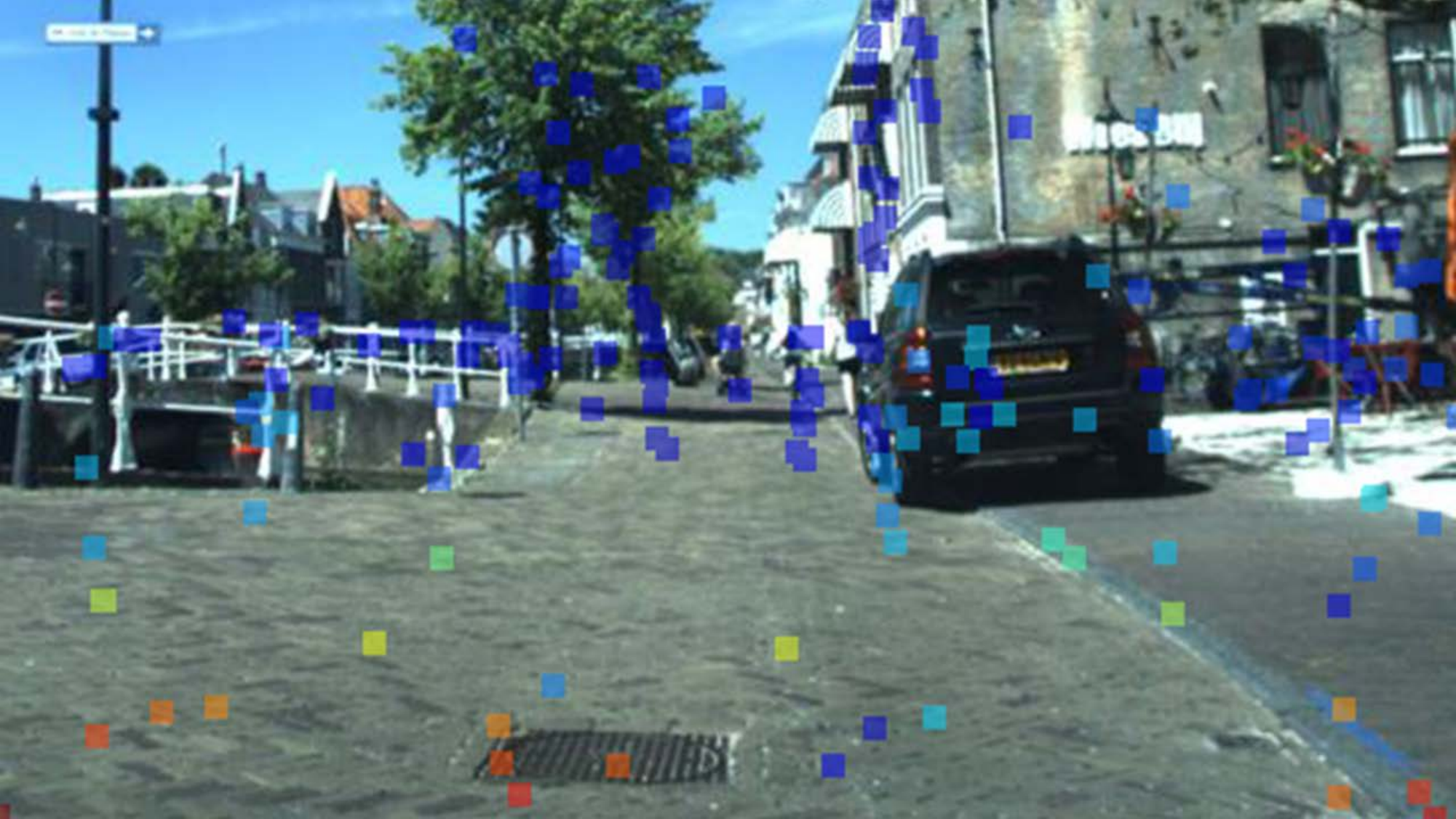}%
\label{fig:scene3 rad pred}}

\subfloat[BEV input]{\includegraphics[width=0.32\textwidth]{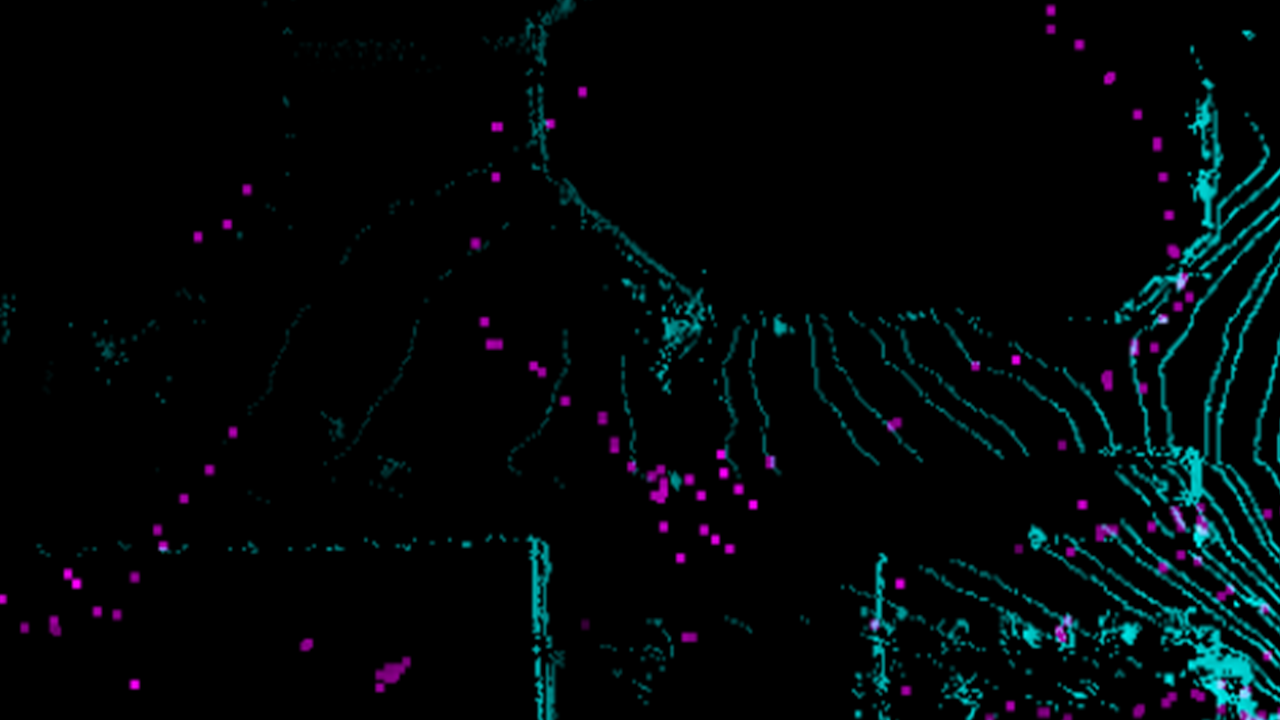}%
\label{fig:scene3 bev in}}
\hfil
\subfloat[BEV ground truth]{\includegraphics[width=0.32\textwidth]{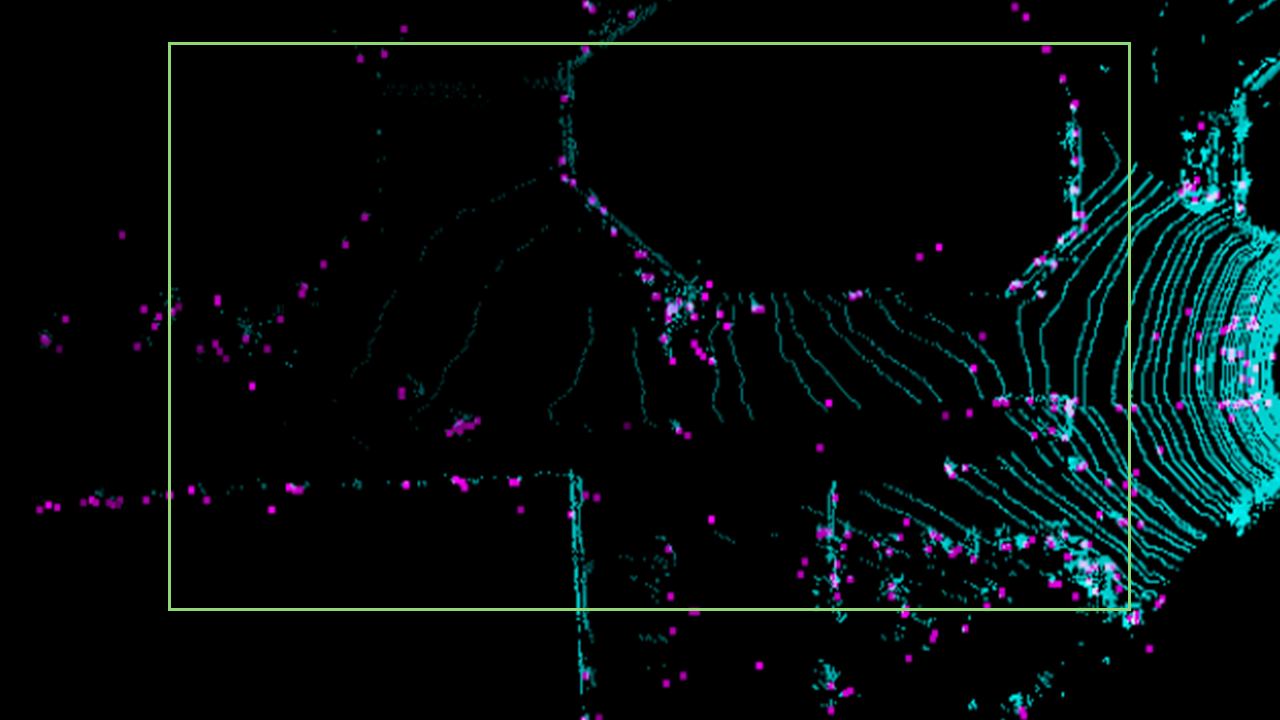}%
\label{fig:scene3 bev gt}}
\hfil
\subfloat[BEV prediction]{\includegraphics[width=0.32\textwidth]{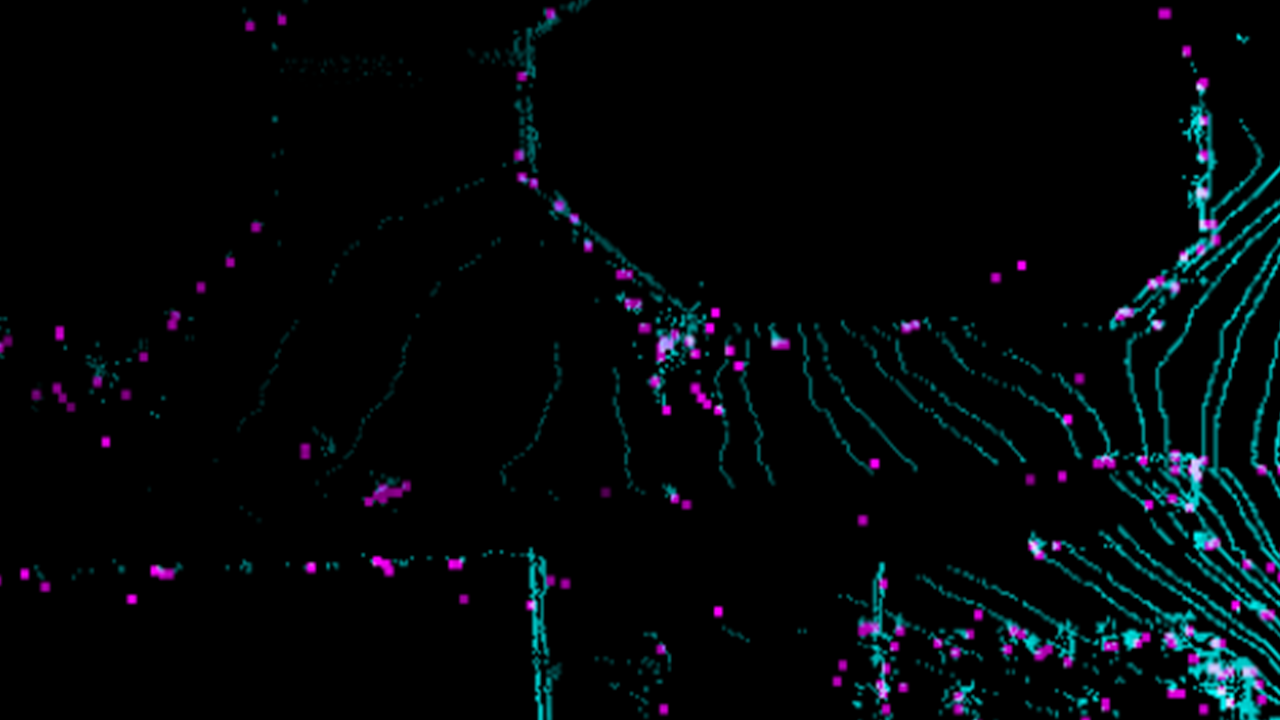}%
\label{fig:scene3 bev pred}}
\caption{Calibration results of RLCNet after iterative refinement (Scene 2)}
\label{fig:rlcnet_results2}
\end{figure*}

Since these random transformations are applied when both LiDAR and RADAR data are in the same coordinate system, the miscalibration between LiDAR and RADAR can be computed using the loop closure constraint, as shown below:

\begin{equation}
    \mathbf{\Delta T}_{RL} = \mathbf{\Delta T}_{RC} \cdot (\mathbf{\Delta T}_{LC})^{-1}.
\end{equation}

An illustration of this process is shown in Figure \ref{fig:deltat}. During training, the generated random transformations serve as ground truth calibration parameters for the network to predict. By randomly varying the deviation for each training iteration, we address the issue of data insufficiency. Additionally, photometric transformations are randomly applied to the RGB image in each iteration to mitigate overfitting.

\subsubsection{\textbf{Training Details}}

The network is trained on an Nvidia RTX 4090 GPU with a batch size of 60. The Adam optimizer is used, with an initial learning rate of $1 \times 10^{-5}$. The first network, which exhibits the largest miscalibration range, is trained for 140 epochs and serves as the pre-trained model for subsequent networks. As these networks are initialized with the pre-trained weights, those with smaller miscalibration ranges are trained for only 60 epochs.

\section{\textbf{Experiments}}\label{sec:experiments}

\begin{table*}
\settowidth\rotheadsize{Translation}
\centering
\renewcommand{\arraystretch}{1.3}
\caption{Mean prediction errors from RLCNet trained on different miscalibration ranges. \emph{Direct} and \emph{Soft} indicate\\the \emph{Feature Sharing Scheme} employed; $E_t$ and $E_r$ indicate the absolute translation and rotation error}
\label{tab:rlcnet_iterative}
\begin{tabular}{c c c | c c | c c | c c | c c | c c}   
\toprule
\multicolumn{3}{c}{\multirow{2}{*}{\makecell{Network\\miscalibration\\ ranges}}} & \multicolumn{2}{|c|}{\makecell{Network\\$10^{\circ}/50cm$}} & \multicolumn{2}{|c|}{\makecell{Network\\$6^{\circ}/30cm$}} & \multicolumn{2}{|c|}{\makecell{Network\\$4^{\circ}/20cm$}} & \multicolumn{2}{|c|}{\makecell{Network\\$2^{\circ}/10cm$}} & \multicolumn{2}{|c}{\makecell{Network\\$1^{\circ}/5cm$}} \\
\cmidrule{4-13}
 &  &  & \textbf{Direct} & \textbf{Soft} & \textbf{Direct} & \textbf{Soft} & \textbf{Direct} & \textbf{Soft} & \textbf{Direct} & \textbf{Soft} & \textbf{Direct} & \textbf{Soft} \\
 \midrule
\parbox[t]{2mm}{\multirow{8}{*}{\rotatebox[origin=c]{90}{\textbf{LiDAR-Camera}}}} & \multirow{4}{*}{\rothead{Rotation \\Error ($^{\circ}$)}} & $E_r$ & 1.294 & \underline{\textbf{1.177}} & 0.799 & \underline{\textbf{0.686}} & 0.470 & \underline{\textbf{0.387}} & \underline{\textbf{0.338}} & 0.370 & 0.244 & \underline{\textbf{0.220}} \\
 &  & Roll & 0.735 & \underline{\textbf{0.606}} & 0.424 & \underline{\textbf{0.331}} & 0.172 & \underline{\textbf{0.170}} & 0.223 & \underline{\textbf{0.199}} & 0.126 & \underline{\textbf{0.097}} \\
 &  & Pitch & 0.597 & \underline{\textbf{0.507}} & \underline{\textbf{0.327}} & 0.341 & 0.198 & \underline{\textbf{0.159}} & 0.191 & \underline{\textbf{0.179}} & 0.099 & \underline{\textbf{0.096}} \\
 &  & Yaw & \underline{\textbf{0.667}} & 0.675 & 0.416 & \underline{\textbf{0.369}} & \underline{\textbf{0.192}} & 0.251 & 0.235 & \underline{\textbf{0.198}} & 0.163 & \underline{\textbf{0.138}} \\
 \cmidrule{2-13}
 & \multirow{4}{*}{\rothead{Translation Error (cm)}} & $E_t$ & 5.949 & \underline{\textbf{4.756}} & 2.593 & \underline{\textbf{2.574}} & 1.820 & \underline{\textbf{1.563}} & 1.574 & \underline{\textbf{1.569}} & 1.048 & \underline{\textbf{0.926}} \\
 &  & X & 3.815 & \underline{\textbf{2.736}} & \underline{\textbf{1.567}} & 1.633 & 1.086 & \underline{\textbf{0.841}} & 1.062 & \underline{\textbf{0.879}} & 0.620 & \underline{\textbf{0.511}} \\
 &  & Y & 2.439 & \underline{\textbf{1.965}} & \underline{\textbf{0.896}} & 1.024 & \underline{\textbf{0.518}} & 0.715 & 0.770 & \underline{\textbf{0.706}} & 0.365 & \underline{\textbf{0.351}} \\
 &  & Z & 2.79 & \underline{\textbf{2.413}} & \underline{\textbf{0.963}} & 1.240 & 0.979 & \underline{\textbf{0.843}} & 0.997 & \underline{\textbf{0.874}} & 0.606 & \underline{\textbf{0.550}} \\
 \midrule
\parbox[t]{2mm}{\multirow{8}{*}{\rotatebox[origin=c]{90}{\textbf{RADAR-Camera}}}} & \multirow{4}{*}{\rothead{Rotation \\Error ($^{\circ}$)}} & $E_r$ & 1.651 & \underline{\textbf{1.471}} & 1.064 & \underline{\textbf{0.870}} & 0.630 & \underline{\textbf{0.563}} & 0.603 & \underline{\textbf{0.496}} & 0.245 & \underline{\textbf{0.232}} \\
 &  & Roll & 0.887 & \underline{\textbf{0.713}} & 0.444 & \underline{\textbf{0.427}} & 0.339 & \underline{\textbf{0.261}} & 0.241 & \underline{\textbf{0.230}} & 0.168 & \underline{\textbf{0.133}} \\
 &  & Pitch & 0.734 & \underline{\textbf{0.572}} & 0.409 & \underline{\textbf{0.350}} & 0.259 & \underline{\textbf{0.248}} & 0.219 & \underline{\textbf{0.204}} & 0.100 & \underline{\textbf{0.090}} \\
 &  & Yaw & 0.972 & \underline{\textbf{0.913}} & 0.716 & \underline{\textbf{0.561}} & 0.424 & \underline{\textbf{0.356}} & \underline{\textbf{0.267}} & 0.314 & 0.134 & \underline{\textbf{0.132}} \\
 \cmidrule{2-13}
 & \multirow{4}{*}{\rothead{Translation Error (cm)}} & $E_t$ & \underline{\textbf{9.138}} & 9.261 & \underline{\textbf{2.954}} & 4.188 & 3.088 & \underline{\textbf{2.798}} & 2.804 & \underline{\textbf{2.658}} & 1.563 & \underline{\textbf{1.430}} \\
 &  & X & \underline{\textbf{5.617}} & 6.243 & 2.982 & \underline{\textbf{2.689}} & 1.597 & \underline{\textbf{1.529}} & 1.390 & \underline{\textbf{1.284}} & 0.839 & \underline{\textbf{0.824}} \\
 &  & Y & 3.207 & \underline{\textbf{2.781}} & 1.761 & \underline{\textbf{1.522}} & \underline{\textbf{1.037}} & 1.050 & 1.444 & \underline{\textbf{1.111}} & 0.909 & \underline{\textbf{0.703}} \\
 &  & Z & 5.187 & \underline{\textbf{4.659}} & 2.695 & \underline{\textbf{2.090}} & 1.723 & \underline{\textbf{1.661}} & \underline{\textbf{1.553}} & 1.671 & 0.878 & \underline{\textbf{0.697}} \\
 \midrule
\parbox[t]{2mm}{\multirow{8}{*}{\rotatebox[origin=c]{90}{\textbf{LiDAR-RADAR}}}} & \multirow{4}{*}{\rothead{Rotation \\Error ($^{\circ}$)}} & $E_r$ & 1.543 & \underline{\textbf{1.519}} & 0.841 & \underline{\textbf{0.781}} & 0.478 & \underline{\textbf{0.401}} & 0.455 & \underline{\textbf{0.400}} & \underline{\textbf{0.239}} & 0.240 \\
 &  & Roll & 0.947 & \underline{\textbf{0.714}} & 0.448 & \underline{\textbf{0.410}} & \underline{\textbf{0.211}} & 0.231 & 0.251 & \underline{\textbf{0.222 }}& 0.158 & \underline{\textbf{0.125}} \\
 &  & Pitch & \underline{\textbf{0.815}} & 0.818 & \underline{\textbf{0.438}} & 0.442 & 0.231 & \underline{\textbf{0.198}} & 0.212 & \underline{\textbf{0.200}} & 0.122 & \underline{\textbf{0.114}} \\
 &  & Yaw & 0.632 & \underline{\textbf{0.589}} & \underline{\textbf{0.321}} & 0.346 & 0.224 & \underline{\textbf{0.189}} & \underline{\textbf{0.156}} & 0.197 & 0.154 & \underline{\textbf{0.135}} \\
 \cmidrule{2-13}
 & \multirow{4}{*}{\rothead{Translation Error (cm)}} & $E_t$ & \underline{\textbf{7.942}} & 8.64 & 5.039 & \underline{\textbf{4.324}} & \underline{\textbf{2.499}} & 2.745 & 2.810 & \underline{\textbf{2.538}} & 1.237 & \underline{\textbf{1.227}} \\
 &  & X & \underline{\textbf{4.488}} & 4.708 & 2.228 & \underline{\textbf{2.218}} & \underline{\textbf{1.156}} & 1.395 & 1.339 & \underline{\textbf{1.219}} & 0.815 & \underline{\textbf{0.709}} \\
 &  & Y & \underline{\textbf{3.163}} & 3.918 & 1.966 & \underline{\textbf{1.799}} & 1.034 & \underline{\textbf{0.993}} & 1.098 & \underline{\textbf{1.053}} & 0.650 & \underline{\textbf{0.525}} \\
 &  & Z & 4.533 & \underline{\textbf{4.495}} & 3.062 & \underline{\textbf{2.621}} & 1.916 & \underline{\textbf{1.732}} & 1.939 & \underline{\textbf{1.589}} & 0.809 & \underline{\textbf{0.672}} \\
 \bottomrule

\end{tabular}
\end{table*}

To evaluate the calibration accuracy, the translational and rotational components are analyzed separately. For translation, the Euclidean distance $E_t$ between the predicted and ground-truth vectors serves as the primary metric, with per-axis errors (X, Y, Z) offering further insight. For rotation, the angular distance $E_r$ between quaternions is computed, and the error is also decomposed into Euler angles (Roll, Pitch, Yaw) for detailed analysis.

Table~\ref{tab:rlcnet_iterative} summarizes the mean prediction errors at each refinement stage of RLCNet and compares performance using direct and soft fusion schemes in the feature sharing module. The results clearly show that soft fusion yields superior accuracy and is therefore adopted as the default configuration. The consistent error reduction across stages demonstrates the effectiveness of the multi-stage refinement strategy.

For the LiDAR-Camera calibration, RLCNet achieves a mean absolute rotation error of $0.220^{\circ}$, with individual mean angular errors broken down as follows: roll: $0.097^{\circ}$, pitch: $0.096^{\circ}$, and yaw: $0.138^{\circ}$. In terms of translation, the mean absolute error is $0.926 cm$, with axis-specific errors: X: $0.511 cm$, Y: $0.351 cm$, and Z: $0.550 cm$.

For the RADAR-Camera calibration, the network achieves a mean absolute rotation error of $0.232^{\circ}$, with mean angular errors of roll: $0.133^{\circ}$, pitch: $0.090^{\circ}$, and yaw: $0.132^{\circ}$. The mean absolute translation error is $1.430 cm$, with the corresponding errors in X: $0.824 cm$, Y: $0.703 cm$, and Z: $0.697 cm$.

For the LiDAR-RADAR calibration, the mean absolute rotation error is $0.240^{\circ}$, with mean angular errors of roll: $0.125^{\circ}$, pitch: $0.114^{\circ}$ and yaw: $0.135^{\circ}$, and the mean absolute translation error is $1.227 cm$, with mean translation errors of  X: $0.709 cm$, Y: $0.525 cm$ and Z: $0.672 cm$.

The images in Figures \ref{fig:rlcnet_results1} and \ref{fig:rlcnet_results2} visually demonstrate the improvements in extrinsic calibration parameters achieved through the iterative refinement process of RLCNet.

\begin{figure*}[!t]
\centering
\subfloat[LiDAR inputs]{\includegraphics[width=0.49\textwidth]{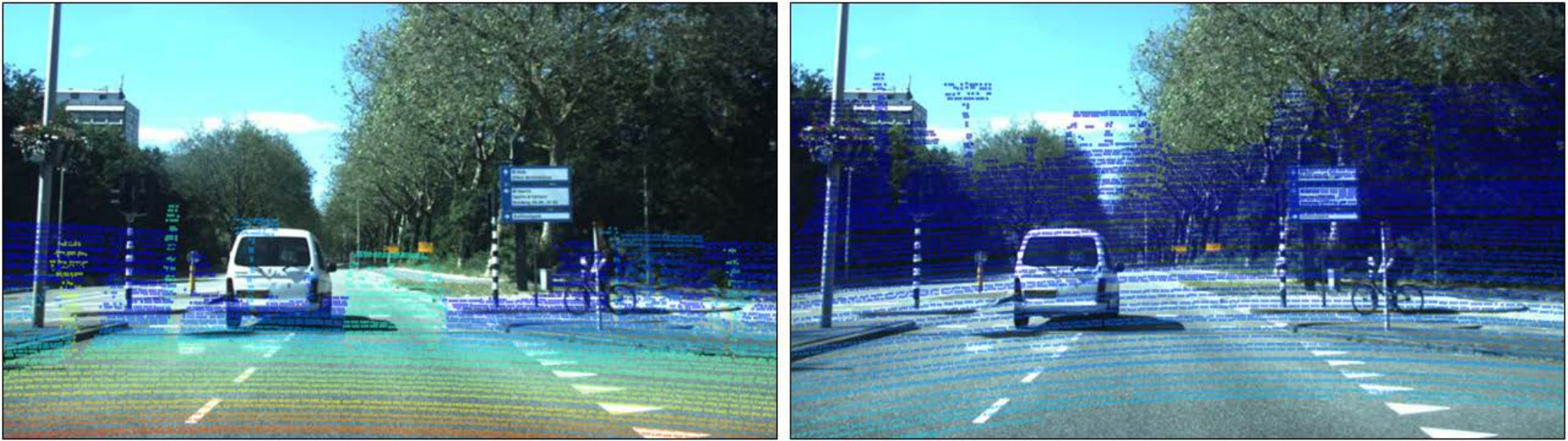}%
\label{fig:rlcnet_repeat_lid_in}}
\hfil
\subfloat[RADAR inputs]{\includegraphics[width=0.49\textwidth]{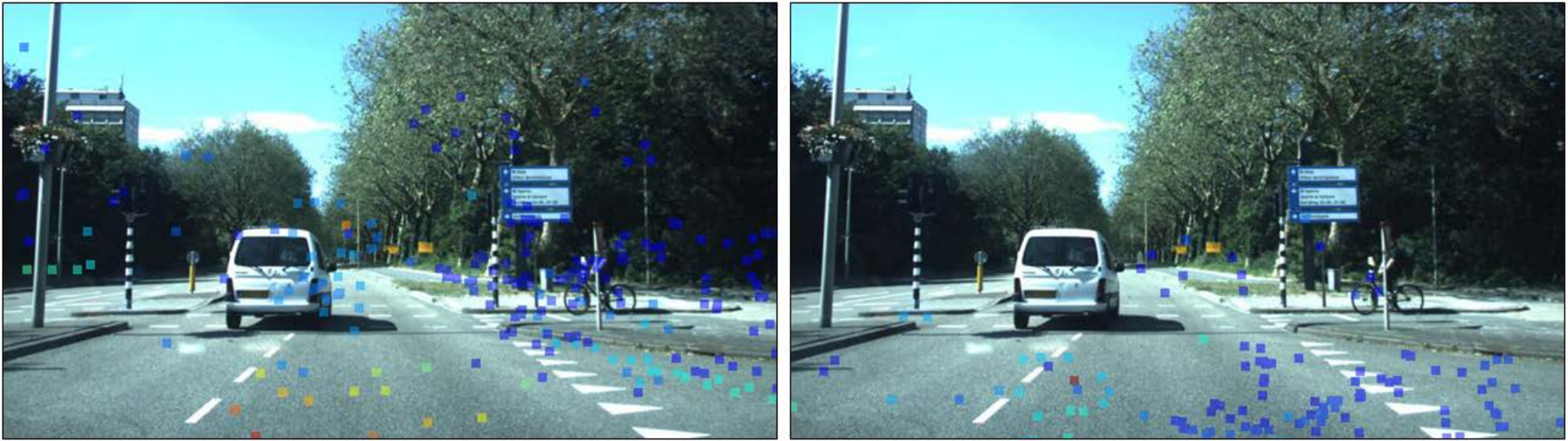}%
\label{fig:rlcnet_repeat_rad_in}}

\subfloat[LiDAR predictions]{\includegraphics[width=0.49\textwidth]{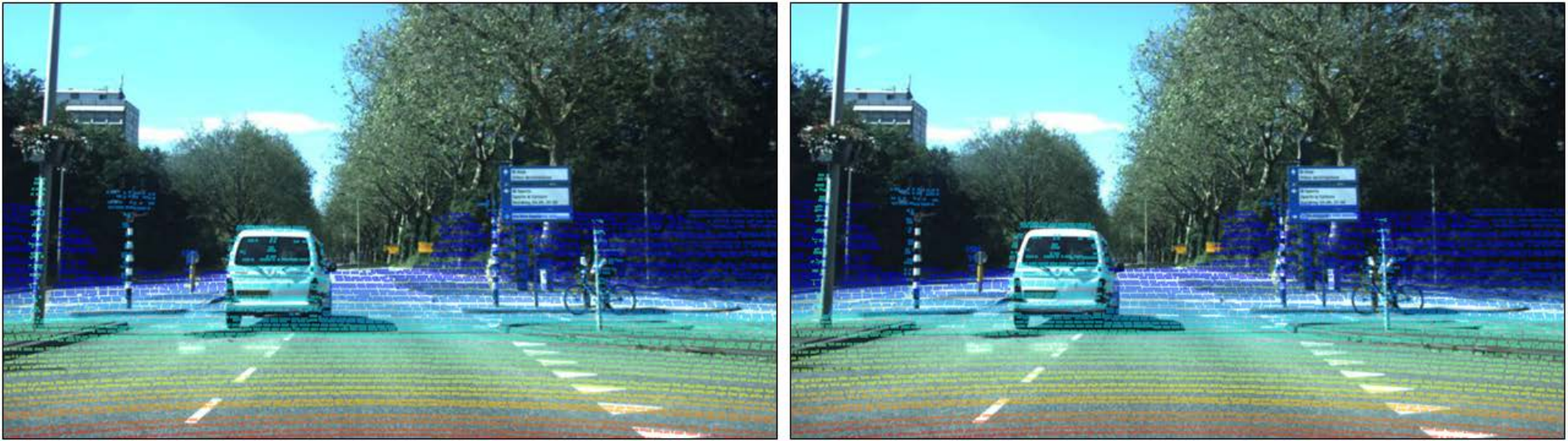}%
\label{fig:rlcnet_repeat_lid_pred}}
\hfill
\subfloat[RADAR predictions]{\includegraphics[width=0.49\textwidth]{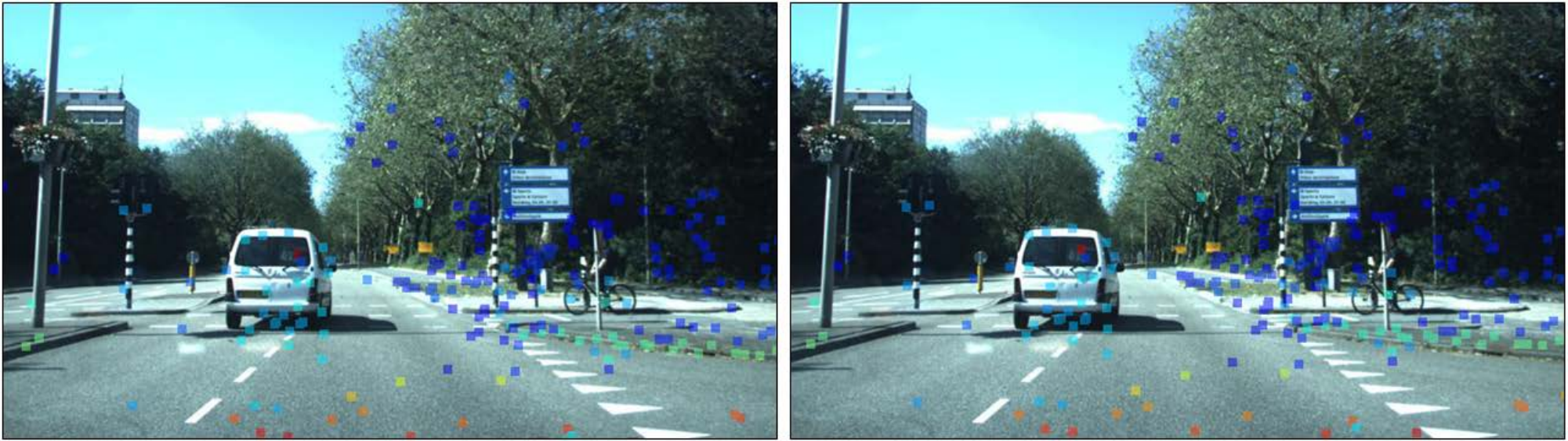}%
\label{fig:rlcnet_repeat_rad_pred}}

\subfloat[Rotation error of Accuracy Analysis]{\includegraphics[width=\textwidth]{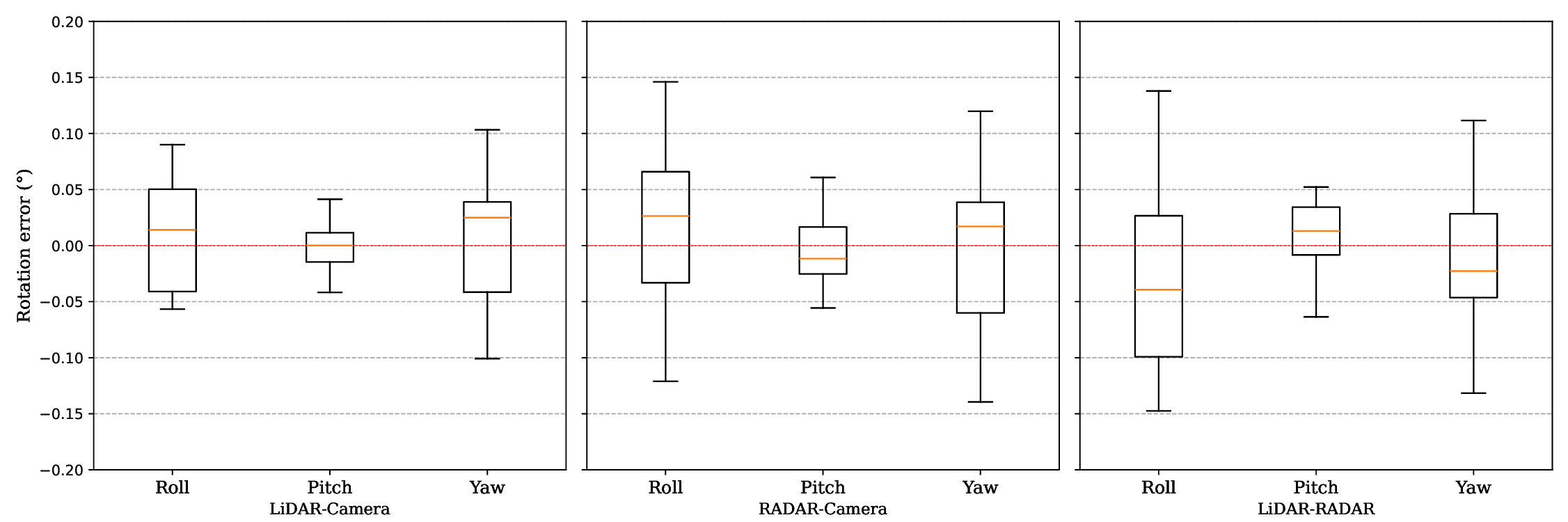}%
\label{fig:rlcnet_accuracy_rot}}

\subfloat[Translation error of Accuracy Analysis]{\includegraphics[width=\textwidth]{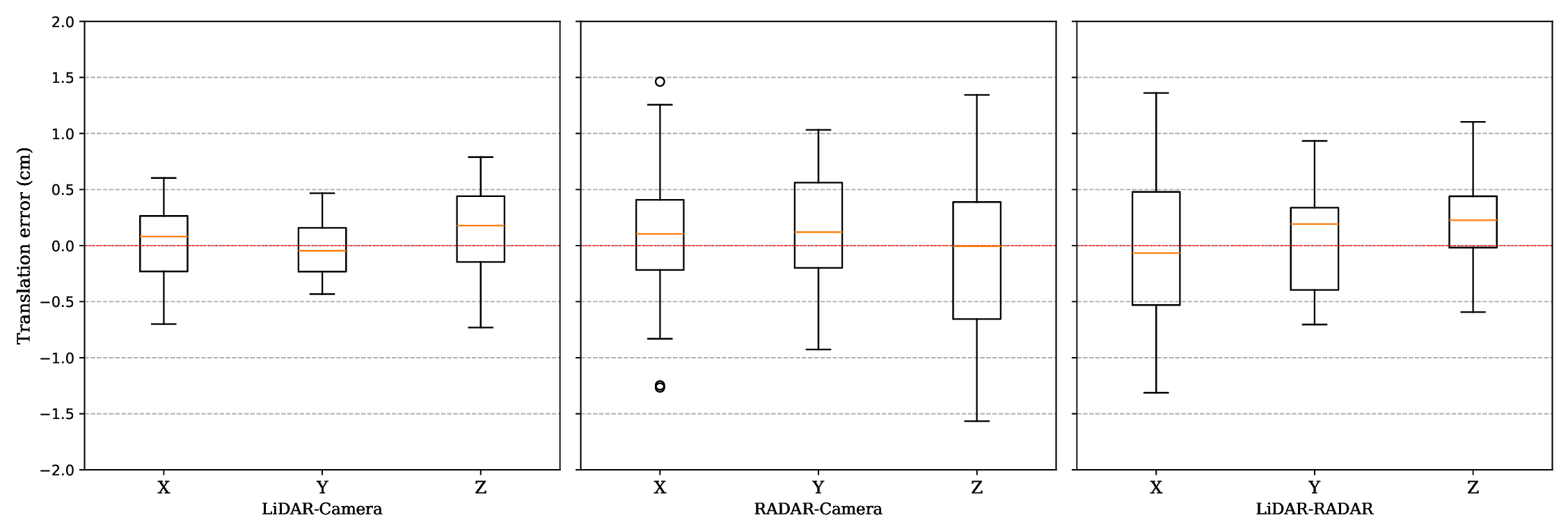}%
\label{fig:rlcnet_accuracy_trans}}
\caption{Results of the Accuracy analysis of RLCNet}
\label{fig:rlcnet_accuracy}
\end{figure*}

\subsection{\textbf{Accuracy Analysis}}
We evaluated the accuracy of RLCNet by introducing varying initial miscalibrations to the same input frame and repeating the test 100 times using a sample from the test set. The resulting prediction error distributions are shown in Figures~\ref{fig:rlcnet_accuracy_rot} and~\ref{fig:rlcnet_accuracy_trans}.

Notably, the pitch angle exhibits the tightest error distribution across all sensor pairs. This can be attributed to the overlapping fields of view and the projection image features, which offer robust cues for pitch estimation. Similarly, translation errors are lowest for the LiDAR-camera pair, likely due to the high-resolution LiDAR point clouds that provide rich geometric details for precise alignment.

Figures~\ref{fig:rlcnet_repeat_lid_in} to~\ref{fig:rlcnet_repeat_rad_pred} visually demonstrate the consistency of RLCNet’s predictions, even under significant variations in initial miscalibration.

\subsection{\textbf{Ablation study}}

\begin{figure*}[!t]
\centering
\subfloat[Rotation error vs $d$]{\includegraphics[width=0.32\textwidth]{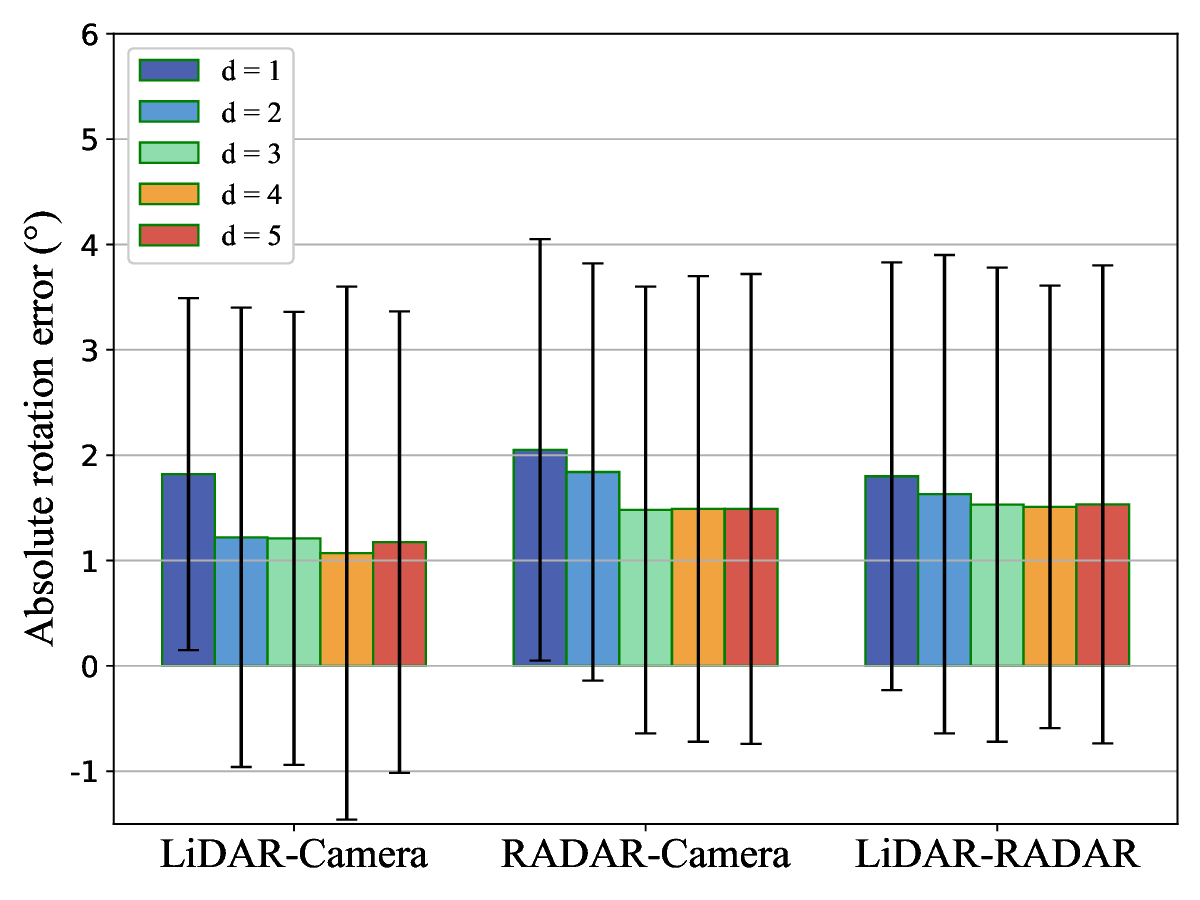}%
\label{fig:rlcnet_ablation_md_rot}}
\hfil
\subfloat[Translation error vs $d$]{\includegraphics[width=0.32\textwidth]{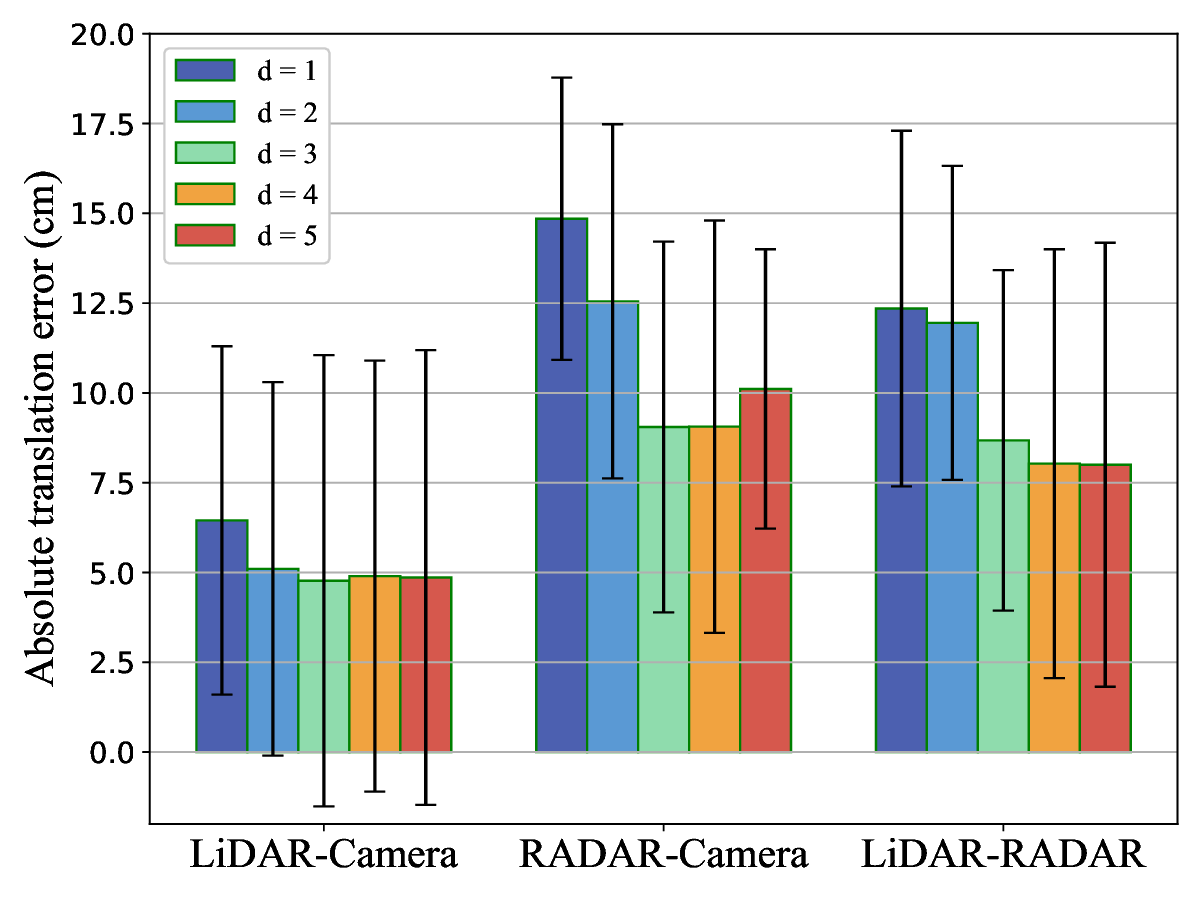}%
\label{fig:rlcnet_ablation_md_trans}}
\hfil
\subfloat[Achievable frequency vs $d$]{\includegraphics[width=0.32\textwidth]{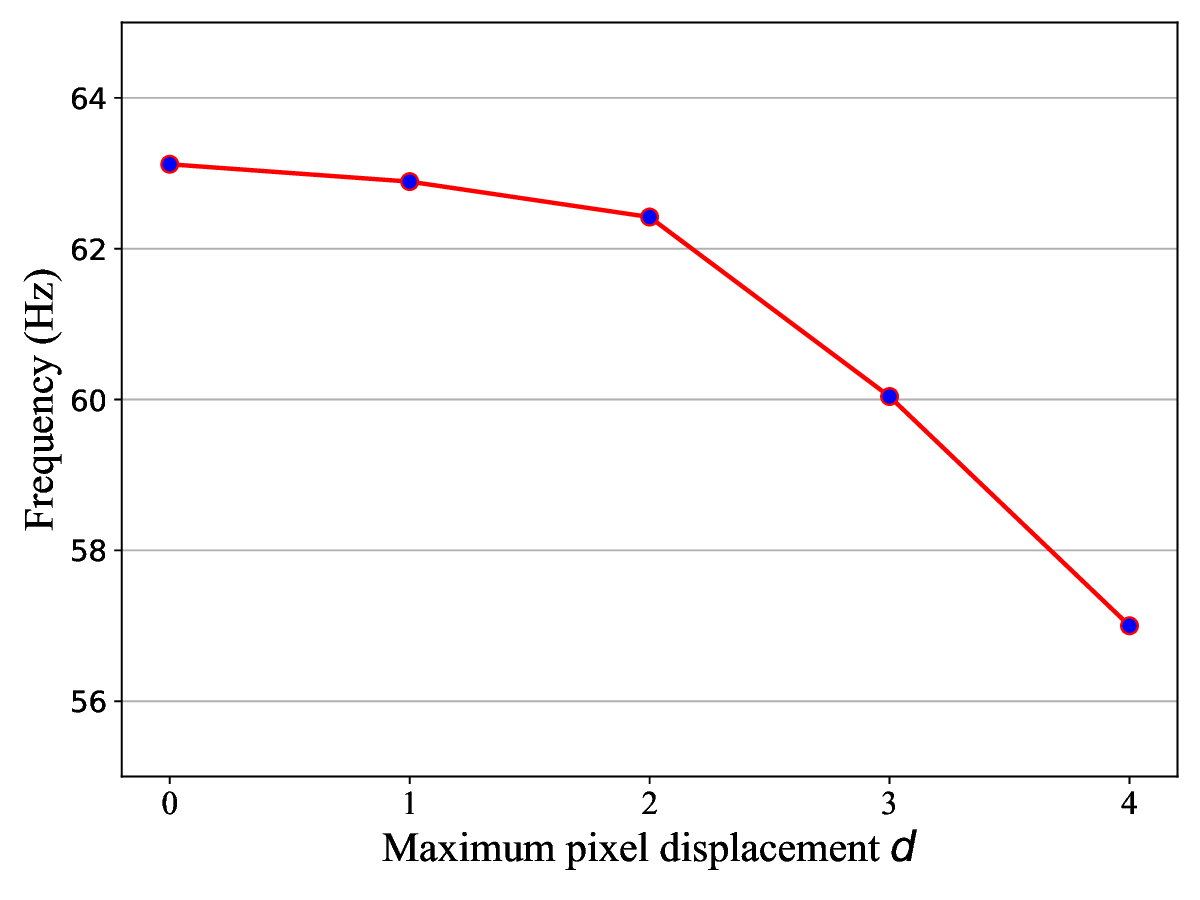}%
\label{fig:rlcnet_ablation_md_line}}

\caption{Results of ablation study by varying maximum distance of\\ correlation layer}
\label{fig:rlcnet_ablation_md}
\end{figure*}

To determine the optimal configuration for RLCNet, we conducted ablation studies on two key parameters: maximum displacement $d$ in the correlation layer and the number of message passing iterations in the MPN.

First, we trained multiple networks with different values of $d$ using an input miscalibration range of $[-10^{\circ}, 10^{\circ}]$ (rotation) and $[-50\,\text{cm}, 50\,\text{cm}]$ (translation). As shown in Figures~\ref{fig:rlcnet_ablation_md_rot} and~\ref{fig:rlcnet_ablation_md_trans}, prediction accuracy improves up to $d=3$, beyond which no significant or consistent gains are observed. Given the exponential increase in computational cost with larger $d$ (Figure~\ref{fig:rlcnet_ablation_md_line}), we select $d=3$ as the optimal value.

\begin{figure*}[!t]
\centering
\subfloat[Rotation error vs mpi]{\includegraphics[width=0.32\textwidth]{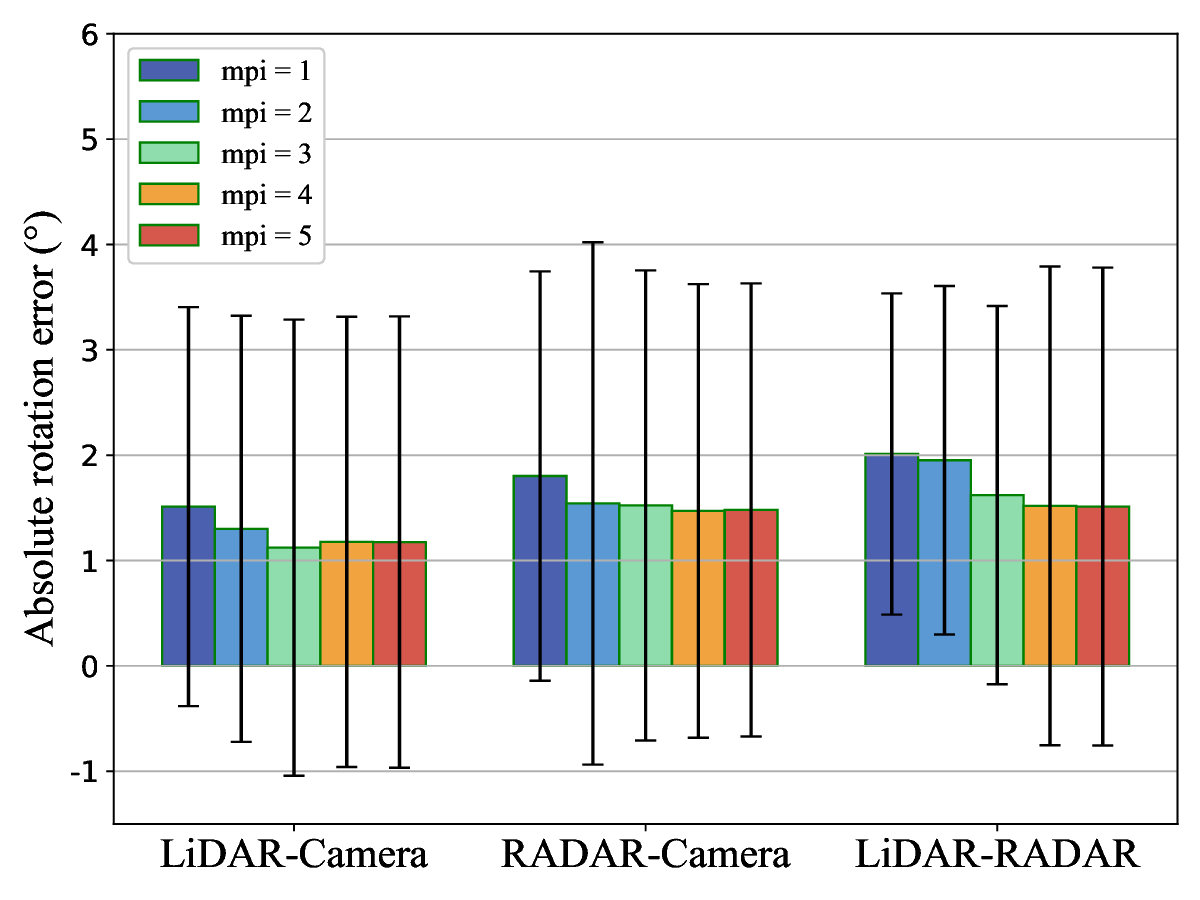}%
\label{fig:rlcnet_ablation_mpn_rot}}
\hfil
\subfloat[Translation error vs mpi]{\includegraphics[width=0.32\textwidth]{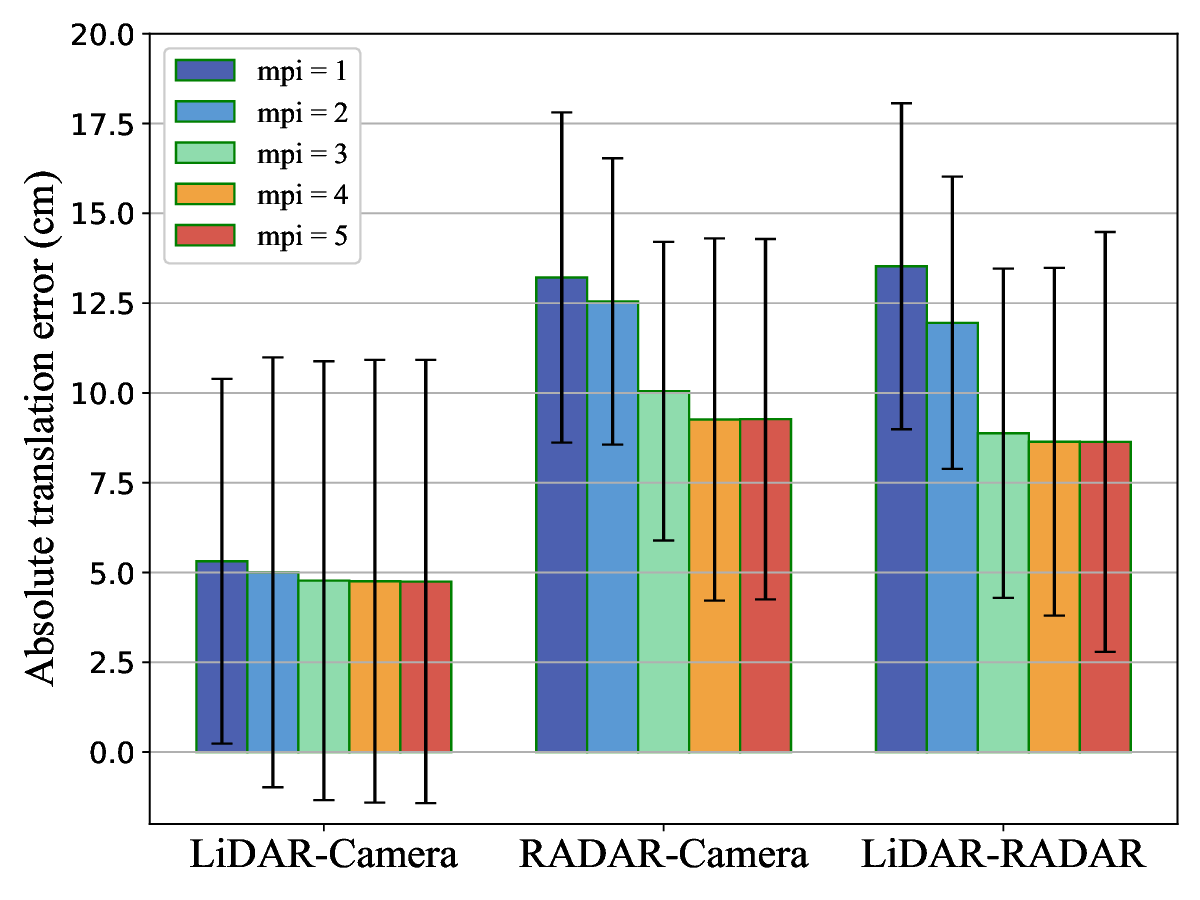}%
\label{fig:rlcnet_ablation_mpn_trans}}
\hfil
\subfloat[Evolution of $\alpha$ in network with mpi $=5$]{\includegraphics[width=0.32\textwidth]{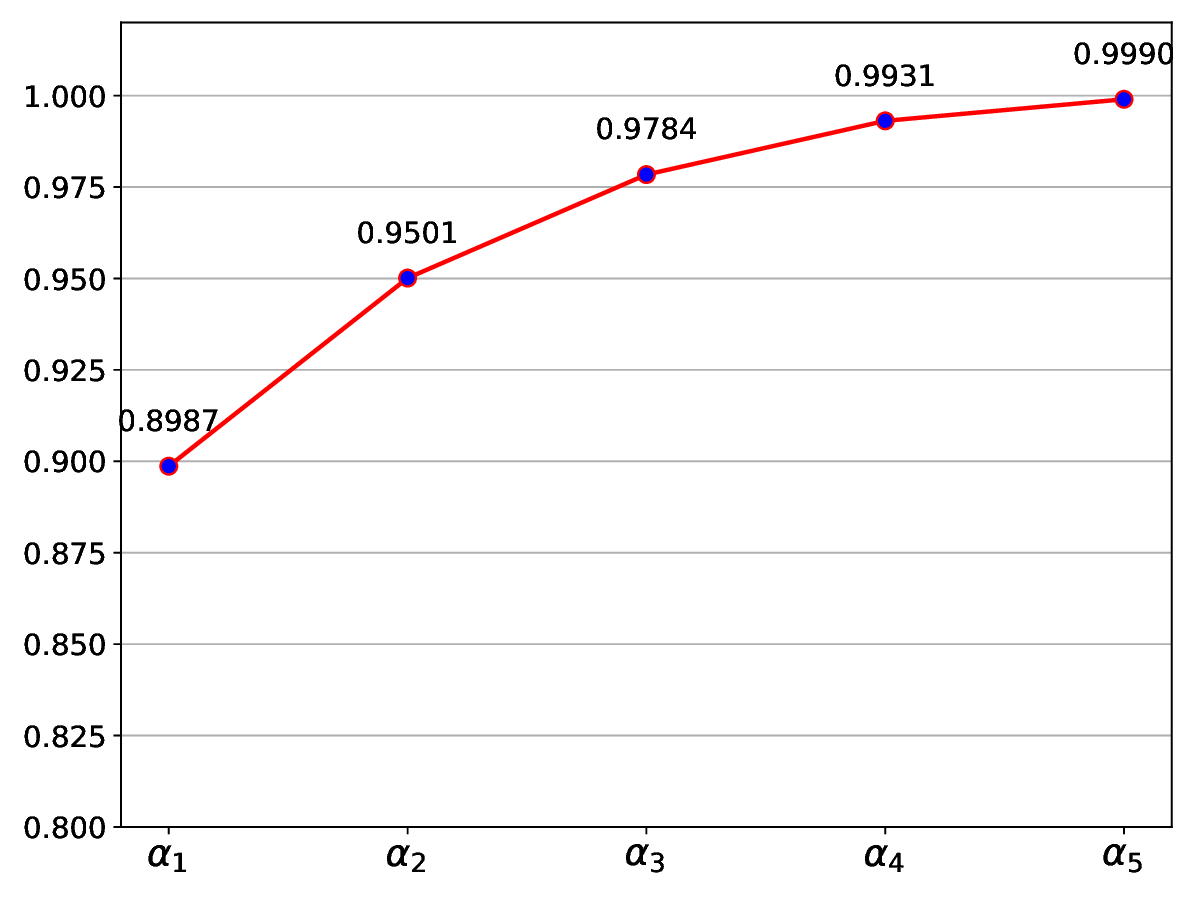}%
\label{fig:rlcnet_ablation_mpn_alpha}}

\caption{Results of ablation study by varying the number of message passing iterations ($mpi$)}
\label{fig:rlcnet_ablation_mpn}
\end{figure*}

Next, we evaluated the impact of message passing iterations by varying the iteration count. Figures~\ref{fig:rlcnet_ablation_mpn_rot} and~\ref{fig:rlcnet_ablation_mpn_trans} show that accuracy improves with more iterations, but plateaus beyond four. Figure~\ref{fig:rlcnet_ablation_mpn_alpha} further illustrates that $\alpha_5$ approaches 1, rendering further updates redundant. Since the MPN cost grows linearly with the number of iterations, we select 4 iterations as the optimal setting.

\subsection{\textbf{Comparison with baselines}}

\begin{table*}[t]
\centering
\renewcommand{\arraystretch}{1.5}
\caption{Comparison of rotational calibration error of RLCNet with joint calibration framework in literature.\\ \emph{Direct} and \emph{Soft} indicate the \emph{Feature Sharing Scheme} employed; All reported values are in degrees.}
\label{tab:rlcnet_benchmark_rot}
\begin{tabular}{c c c c c c c c c c} 
\toprule
\multirow{2}{*}{\textbf{Method}} & \multicolumn{3}{c}{\textbf{LiDAR-Camera}} & \multicolumn{3}{c}{\textbf{RADAR-Camera}} & \multicolumn{3}{c}{\textbf{LiDAR-RADAR}} \\
 & \textbf{Roll} & \textbf{Pitch} & \textbf{Yaw} & \textbf{Roll} & \textbf{Pitch} & \textbf{Yaw} & \textbf{Roll} & \textbf{Pitch} & \textbf{Yaw} \\
 \midrule
Per{\v{s}}i{\'c}\cite{pervsic2021online} & \textbf{0.02} & 0.10 & 0.24 & \_ & \_ & 0.26 & \_ & \_ & \textbf{0.03 }\\
RLCNet (Direct) & 0.120 & 0.119 & 0.196 & \textbf{0.170 }& 0.101 & 0.221 & 0.193 & 0.100 & 0.256 \\
RLCNet (Soft) & 0.108 & \textbf{0.096} & \textbf{0.184} & 0.172 &\textbf{ 0.101} & \textbf{0.199} & \textbf{0.181} & \textbf{0.098} & 0.223 \\
\bottomrule

\end{tabular}

\end{table*}

\begin{table*}[t]
\centering
\renewcommand{\arraystretch}{1.5}
\caption{Comparison of pairwise calibration error of RLCNet with joint calibration framework in literature. \emph{Direct} and \emph{Soft} indicate the \emph{Feature Sharing Scheme} employed; \textit{R} indicate the range and \textit{Az} and \textit{El} indicate azimuth and elevation angles.}
\label{tab:rlcnet_benchmark}
\begin{tabular}{c c c c| c c c| c c c| c c c}
\toprule
\multirow{2}{*}{} & \multicolumn{9}{c}{\textbf{Pairwise calibration error}} & \multicolumn{3}{|c}{\textbf{Closed loop error}} \\
 \cmidrule{2-13}
 & \multicolumn{3}{c|}{\textbf{LiDAR- Camera}} & \multicolumn{3}{c|}{\textbf{RADAR-Camera}} & \multicolumn{3}{c|}{\textbf{LiDAR-RADAR}} & \multicolumn{3}{c}{\textbf{Camera-LiDAR-RADAR}} \\
 \cmidrule{2-13}
 & \textbf{Az ($^{\circ}$)} & \textbf{El ($^{\circ}$)} & \textbf{R (cm)} & \textbf{Az ($^{\circ}$)} & \textbf{El ($^{\circ}$)} & \textbf{R (cm)} & \textbf{Az ($^{\circ}$)} & \textbf{El ($^{\circ}$)} & \textbf{R (cm)} & \textbf{Az ($^{\circ}$)} & \textbf{El ($^{\circ}$)} & \textbf{R (cm)} \\
 \midrule
Hayoun\cite{hayoun2024physics} & \textbf{0.08} & 0.20 & \_ & \textbf{0.02} & 0.14 & \_ & 0.51 & 0.61 & 13 & 0.02 & 0.02 & 1 \\
RLCNet (Direct) & 0.179 & 0.039 & 0.520 & 0.101 & 0.091 & 0.752 & 0.159 & 0.039 & 0.534 & 0.022 & 0.017 & 0.431 \\
RLCNet (Soft) & 0.140 & \textbf{0.037} & \textbf{0.415} & 0.144 & \textbf{0.080} & \textbf{0.713} & \textbf{0.153} & \textbf{0.033} & \textbf{0.524 }& \textbf{0.019} & \textbf{0.015} & \textbf{0.378} \\
\bottomrule
\end{tabular}

\end{table*}
We compared RLCNet's calibration accuracy against two targetless joint calibration methods described in Section~\ref{sec:literature}. The method by Per{\v{s}}i{\'c} et al.~\cite{pervsic2021online}, which estimates only rotational parameters, was tested on the NuScenes dataset~\cite{nuscenes}. To ensure a fair comparison, we retrained RLCNet's five iterative networks on the same dataset, closely replicating their experimental setup.

Table~\ref{tab:rlcnet_benchmark_rot} presents results on \textit{scene 343} of the NuScenes dataset (20 seconds duration), where the baseline focuses solely on yaw for RADAR due to the absence of elevation data, limiting correction in roll and pitch. In contrast, RLCNet estimates full 3-DoF rotations and demonstrates lower mean angular error across the scene, with each frame initialized within $[-10^{\circ}, 10^{\circ}]$. RLCNet refines predictions within 54\,ms\footnote{The test was conducted on a consumer-grade laptop equipped with an NVIDIA RTX 3080 GPU.} using five networks, outperforming the baseline, which requires over 2 seconds to reach comparable accuracy.

The second method, by Hayoun et al.~\cite{hayoun2024physics}, evaluates calibration via reprojection error (azimuth, elevation, and range) of manually annotated reflective trihedral corners. As this testbed could not be replicated, we compared RLCNet's results to their reported values. We converted RLCNet's prediction errors into azimuth ($\delta\theta$), elevation ($\delta\phi$), and range ($\delta r$) using the predicted and ground-truth transformations.

As shown in Table~\ref{tab:rlcnet_benchmark}, RLCNet achieves lower elevation errors for all sensor pairs. While the baseline performs slightly better in azimuth for LiDAR-camera and RADAR-camera, RLCNet surpasses it across all metrics for LiDAR-RADAR calibration. Additionally, RLCNet shows significantly lower closed-loop error, highlighting the strength of the MPN-based optimization in ensuring consistent multi-sensor calibration.

\section{\textbf{Online Calibration}}\label{sec:online}

Although RLCNet generates temporally independent predictions from individual frames, it is tailored for online deployment. To exploit temporal consistency in dynamic environments, we introduce an online calibration framework that integrates a weighted moving average and outlier rejection, emphasizing recent predictions. This design reduces sensitivity to noise while remaining responsive to changes in extrinsic parameters. Rotational and translational components are handled separately within both the averaging and outlier detection processes to ensure effective adaptation.

\subsection{\textbf{Weighted Moving Average}}
To define the weighted moving average approach, we use the following notations:
\begin{itemize}
    \item $\mathbf{r}_t \in \mathbb{R}^4$ is the predicted rotational quaternion and $\mathbf{t}_t \in \mathbb{R}^3$ is the predicted translation vector at time $t$
    \item $\Hat{\mathbf{r}}_t$ is the moving average of quaternion and $\Hat{\mathbf{t}}_t$ is the moving average of translation vector at time $t$
    \item $\mathbf{r}_{t-k}$ and $\mathbf{t}_{t-k}$ are the quaternion prediction and translation vector prediction from $k$ time steps ago
    \item $w_k = \alpha^{k}$ is the weight assigned to the prediction $k$ steps ago, where $0 < \alpha < 1$, placing more weight on recent predictions.
    \item $N$ is the length of the moving average window
\end{itemize}
The weights are normalized using:
\begin{equation}
    w'_k = \dfrac{w_k}{\sum_{k=0}^{N-1} w_k}
\end{equation}

\subsubsection{\textbf{Translation vector moving average}}

Since the translation vector is in Euclidean space, a weighted moving average can be directly applied. The estimated translation error is given by:
\begin{equation}
    \Hat{\mathbf{t}}_t = \sum_{k=0}^{N-1} w'_k \mathbf{t}_{t-k}
\end{equation}

\subsubsection{\textbf{Quaternion moving average}}
Since quaternions are not additive in the same way vectors are, we use Spherical Linear Interpolation (SLERP), which is a technique used to interpolate between two unit quaternions. The formula for SLERP between two quaternions $\mathbf{r}_0$ and $\mathbf{r}_1$, with interpolation parameter $p$ (where $p=0$ corresponds to $\mathbf{r}_0$ and $p=1$ corresponds to $\mathbf{r}_1$), is given by:
\begin{equation}
    slerp(\mathbf{r}_0, \mathbf{r}_1, p) = \dfrac{sin((1-p)\theta)}{sin(\theta)} \mathbf{r}_0 + \dfrac{sin(p\theta)}{sin(\theta)} \mathbf{r}_1
\end{equation}
where $\theta = arccos (\langle \mathbf{r}_0, \mathbf{r}_1 \rangle)$ is the angle between the two quaternions.

As the two quaternions become closer, the angle between them becomes smaller, and this formula reduces to the corresponding symmetric formula of linear interpolation:
\begin{equation}
    \lim_{\theta \to 0} slerp(\mathbf{r}_0, \mathbf{r}_1, p) = (1-p) \hspace{2pt}\mathbf{r}_0 + p \hspace{2pt}\mathbf{r}_1
\end{equation}

We extend SLERP to compute the weighted interpolation over a set of $N$ quaternion predictions from time steps $t-(N-1)$ to $t$, with corresponding weights $w'_k$. The weighted SLERP is computed iteratively by applying SLERP between consecutive quaternions over the window. The procedure is as follows:
\begin{itemize}
    \item Start with the most recent quaternion $\mathbf{r}_t$ and interpolate between $\mathbf{r}_t$ and $\mathbf{r}_{t-1}$ using SLERP with weight $w'_1$.
    \item Interpolate the result with $\mathbf{r}_{t-2}$ using SLERP with weight $w'_2$, and so on, until all the quaternions in the window is included.
\end{itemize}
Mathematically, the moving average is given by:
\begin{equation}
\begin{split}
    \Hat{\mathbf{r}}_t = \text{slerp}(\cdots\, \text{slerp}(\text{slerp}(\mathbf{r}_t, \mathbf{r}_{t-1}, w'_1), 
    \mathbf{r}_{t-2}, w'_2), \\
    \dots, \mathbf{r}_{t-N+1}, w'_{N-1})
\end{split}
\end{equation}

\subsection{\textbf{Outlier Detection Based on Consecutive Predictions}}

RLCNet's outlier detection framework performs separate consistency checks on translational and rotational calibration errors. A prediction is deemed valid and included in the moving window only if both components pass their respective checks.

\begin{figure*}[!t]
\centering
\subfloat[Phase 1]{\includegraphics[width=0.32\textwidth]{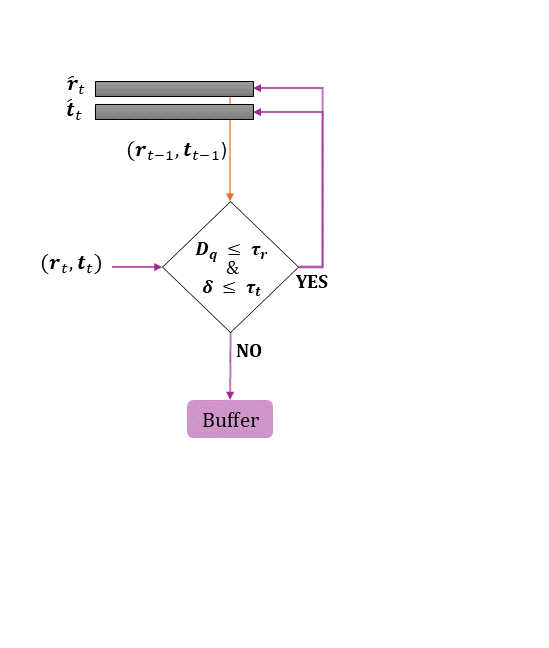}%
\label{fig:rlcnet outlier 1}}
\hfil
\subfloat[Phase 2]{\includegraphics[width=0.32\textwidth]{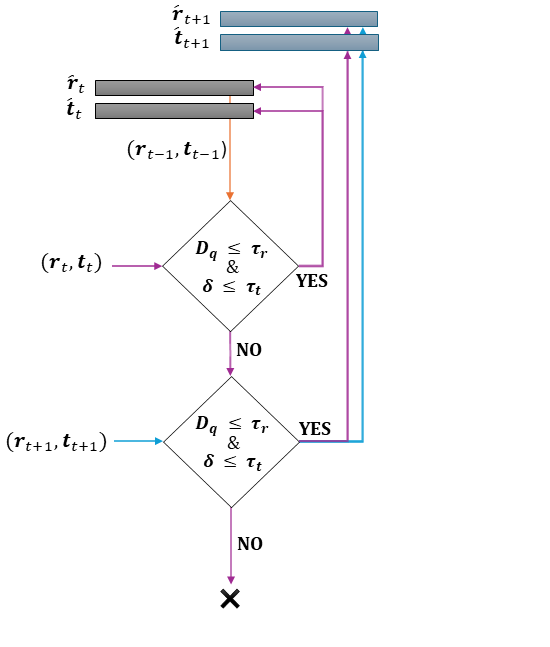}%
\label{fig:rlcnet outlier 2}}
\hfil
\subfloat[Phase 3]{\includegraphics[width=0.32\textwidth]{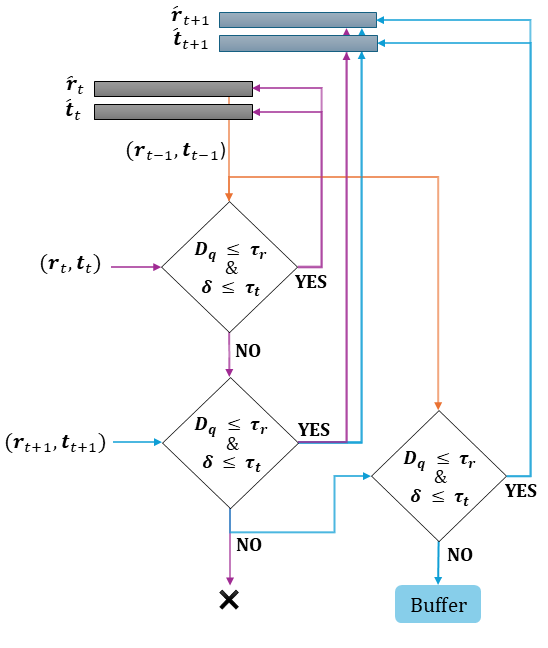}%
\label{fig:rlcnet outlier 3}}

\caption{Flow chart illustrating the outlier detection approach of RLCNet}
\label{fig:rlcnet_outlier_detection}
\end{figure*}

The framework operates in three phases:

\begin{itemize}
    \item \textbf{Phase 1:} Rotational consistency at time step $t$ is verified by computing the angular distance between the current prediction and the latest quaternion included in the moving window, $D_q(\mathbf{r}_t, \mathbf{r}_{t-1})$. The rotation is considered consistent if $D_q(\mathbf{r}_t, \mathbf{r}_{t-1}) \leq \tau_r$.
    
    For translation, we compute the Euclidean distance $\delta(\mathbf{t}_t, \mathbf{t}_{t-1}) = \left\lVert \mathbf{t}_t - \mathbf{t}_{t-1} \right\rVert_2$, and consider it consistent if $\delta(\mathbf{t}_t, \mathbf{t}_{t-1}) \leq \tau_t$.

    If both checks pass, the prediction $(\mathbf{r}_t, \mathbf{t}_t)$ is added to the moving window for averaging. Otherwise, it is stored in a buffer for future comparison.
    \item \textbf{Phase 2:} At time $t{+}1$, the prediction $(\mathbf{r}_{t+1}, \mathbf{t}_{t+1})$ is compared with the buffered prediction $(\mathbf{r}_t, \mathbf{t}_t)$. If both consistency checks are satisfied, that is, $D_q(\mathbf{r}_{t+1}, \mathbf{r}_t) \leq \tau_r$ and $\delta(\mathbf{t}_{t+1}, \mathbf{t}_t) \leq \tau_t$, both the predictions at $t$ and $t{+}1$ are treated as inliers and used to compute updated moving averages $\Hat{\mathbf{r}}_{t+1}$ and $\Hat{\mathbf{t}}_{t+1}$. If either check fails, the buffer is cleared, and the previous prediction is marked as an outlier.

    \item \textbf{Phase 3:} After discarding outliers, the prediction at $t{+}1$ is validated for consistency with the latest entry in the moving window. If both rotation and translation pass the checks, $(\mathbf{r}_{t+1}, \mathbf{t}_{t+1})$ is incorporated into the moving averages $\Hat{\mathbf{r}}_{t+1}$ and $\Hat{\mathbf{t}}_{t+1}$. Otherwise, it is buffered, and the process continues to the next time step. The complete outlier detection process is illustrated in Figure~\ref{fig:rlcnet_outlier_detection}.

\end{itemize}

\subsection{\textbf{Calibration Update}}

At each time step, we compute the weighted average of the predicted rotational and translational errors for the three sensor pairs. A significant change in any of the error estimation indicates a potential shift in the extrinsic parameters. To guide the calibration update, we define two thresholds: $\tau_{cal}^r$ for rotational error and $\tau_{cal}^t$ for translational error. If the estimated rotational or translational error exceeds its respective threshold, the system updates the initial calibration parameters.

\begin{figure}[!t]
\centering
\subfloat[Rotation Noise]{\includegraphics[width=0.49\linewidth]{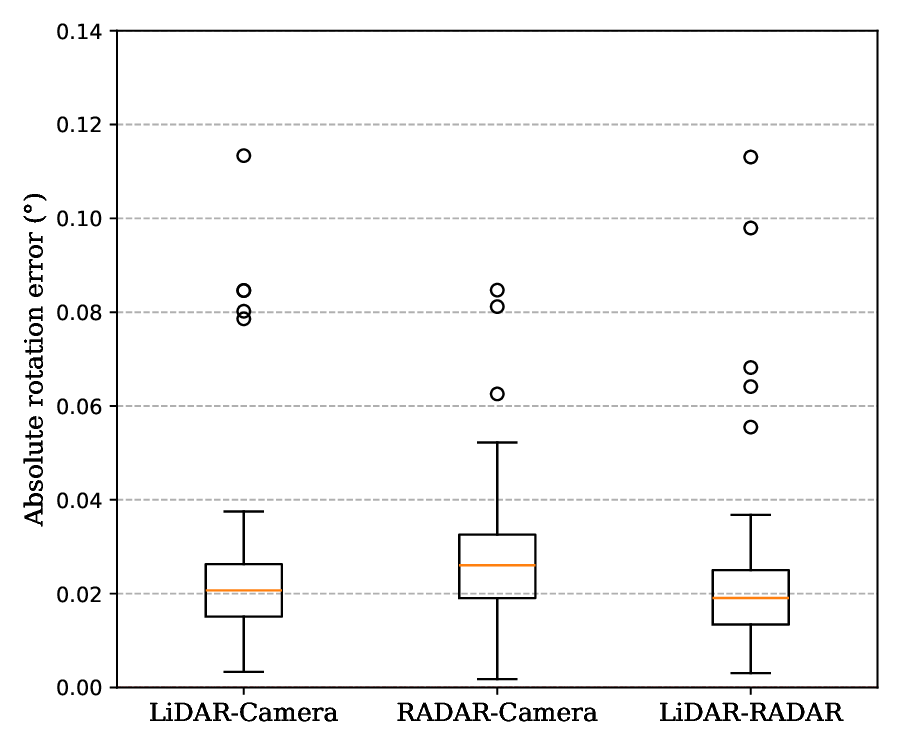}%
\label{fig:rlcnet_rot_noise}}
\hfil
\subfloat[Translation Noise]{\includegraphics[width=0.49\linewidth]{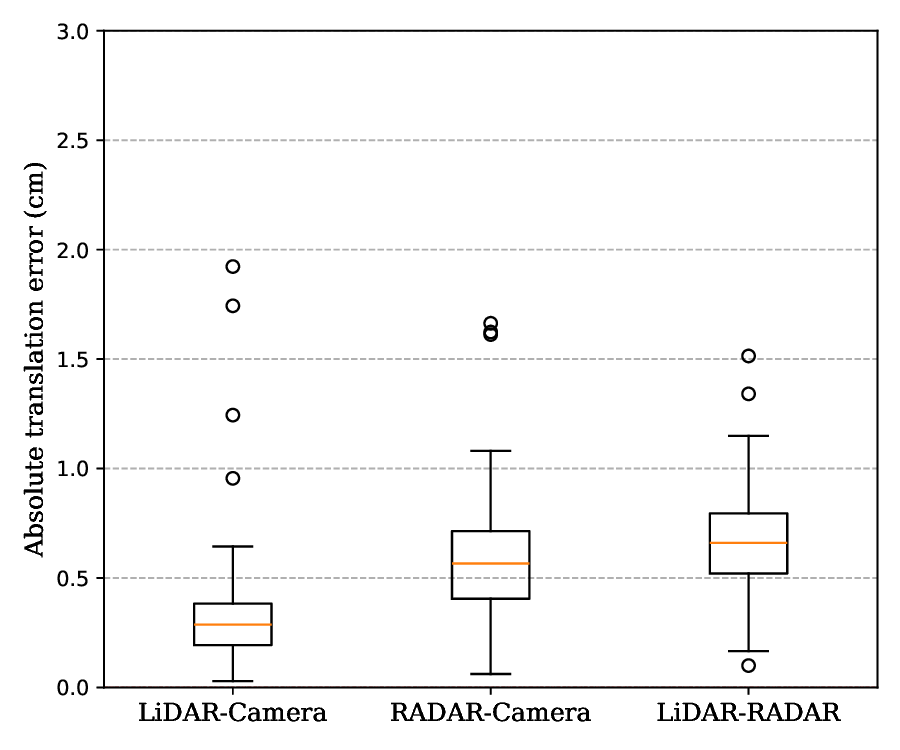}%
\label{fig:rlcnet+_trans_noise}}
\caption{Prediction noise of RLCNet}
\label{fig:rlcnet_noise}
\end{figure}

However, updated calibration parameters must also adhere to the loop closure constraint, ensuring consistency across the sensor network. When a change is detected in one sensor pair, we compare errors from the other two pairs to localize the miscalibrated sensor.

If only one of the two error components, rotation or translation, exceeds its threshold, the decision to retain or update is made by comparing the same component across the other two prediction branches. If both errors surpass thresholds, rotation is prioritized due to its lower noise in RLCNet predictions, as shown in Figure~\ref{fig:rlcnet_noise}.

The RLCNet calibration update framework is summarized as follows:
\begin{itemize}
    \item If $\hspace{4pt}D_q(\Hat{\mathbf{r}}_{t}(x), \mathbf{i}_q) < \tau_{cal}^r \hspace{2pt}\forall \hspace{2pt}x \in [a, b, c]$ where $a$, $b$ and $c$ represent different sensor pairs:
\end{itemize}
\begin{align*}
    \delta(\Hat{\mathbf{t}}_{t}(a), \mathbf{0}_3) \geq \tau_{cal}^t &\Rightarrow \mathbf{T}_{t+1}^{init}(a) = \Hat{\mathbf{T}}_{t}(a) \cdot \mathbf{T}_t^{init}(a)\\[7pt]
    \delta(\Hat{\mathbf{t}}_{t}(b), \mathbf{0}_3) < \delta(\Hat{\mathbf{t}}_{t}(c), \mathbf{0}_3) &\Rightarrow \mathbf{T}_{t+1}^{init}(b) = \mathbf{T}_t^{init}(b)\\[7pt]
    &\Rightarrow \mathbf{T}_{t+1}^{init}(c) = \Acute{\mathbf{T}}_{t}(c) \cdot \mathbf{T}_t^{init}(c)
\end{align*}
\begin{itemize}
    \item Else:
\end{itemize}  
\begin{align*}
    D_q(\Hat{\mathbf{r}}_{t}(a), \mathbf{i}_q) \geq \tau_{cal}^r &\Rightarrow \mathbf{T}_{t+1}^{init}(a) = \Hat{\mathbf{T}}_{t}(a) \cdot \mathbf{T}_t^{init}(a)\\[7pt]
    D_q(\Hat{\mathbf{r}}_{t}(b), \mathbf{i}_q) < D_q(\Hat{\mathbf{r}}_{t}(c), \mathbf{i}_q) &\Rightarrow \mathbf{T}_{t+1}^{init}(b) = \mathbf{T}_t^{init}(b)\\[7pt]
    &\Rightarrow \mathbf{T}_{t+1}^{init}(c) = \Acute{\mathbf{T}}_{t}(c) \cdot \mathbf{T}_t^{init}(c)
\end{align*}

Here, $\mathbf{i}_q$ is the identity quaternion and $\mathbf{0}_3$ is the zero vector. $\Hat{\mathbf{T}}_t$ is the transformation matrix formed by the estimations $\Hat{\mathbf{r}}_{t}$ and $\Hat{\mathbf{t}}_{t}$, while $\Acute{\mathbf{T}}_t$ is derived from the loop closure constraint using predictions from the other sensor pairs.

Once the calibration parameters are updated, the moving windows that estimate the rotation and translation errors are reset and populated with identity quaternions and zero vectors, respectively.

\subsection{\textbf{Choice of parameters}}

To set appropriate thresholds for rotational and translational noise, we analyzed RLCNet's prediction noise in both domains. Based on the distributions shown in Figure~\ref{fig:rlcnet_noise}, we set the rotational threshold to $\tau_r = 0.05^{\circ}$ and the translational threshold to $\tau_t = 1.0$\,cm.

The recalibration thresholds, $\tau_{cal}^r = 0.05^{\circ}$ and $\tau_{cal}^t = 1.0$\,cm, define the maximum allowable miscalibration before triggering correction, reflecting the acceptable limits for rotational and translational deviations.

\begin{figure}
    \centering
    \includegraphics[width=\linewidth]{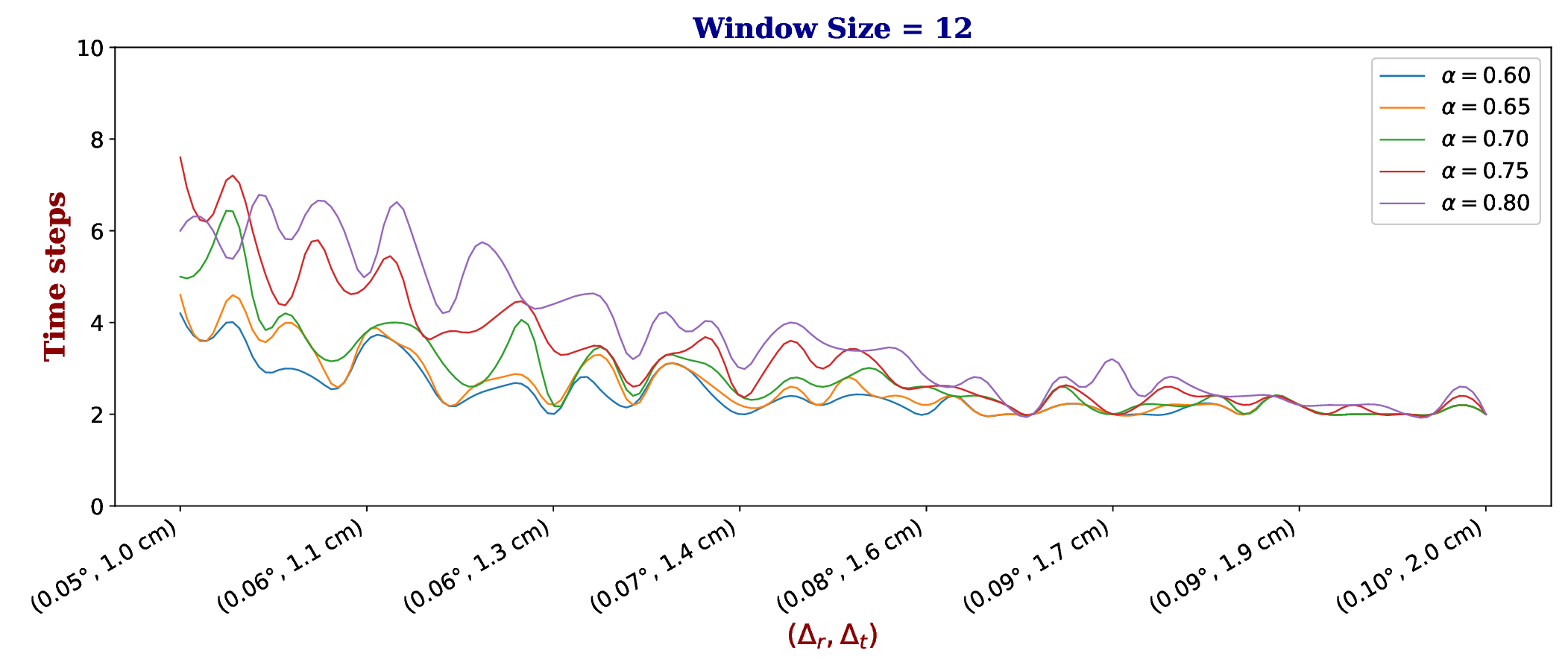}
    \caption{Required time steps to detect changes in extrinsic by RLCNet for different miscalibration ranges}
    \label{fig:rlcnet_window_12}
\end{figure}

Experiments show that wider windows and larger decay factors improve noise filtering, while narrower windows and smaller decay factors enable faster detection of miscalibration. Balancing this trade-off, we selected a window size of $N = 12$ and a decay factor of $\alpha = 0.65$. As shown in Figure~\ref{fig:rlcnet_window_12}, this configuration reliably detects rotational miscalibrations above $0.08^{\circ}$ and translational shifts over $1.6$\,cm within two iterations, demonstrating robust performance in dynamic conditions.

\section{\textbf{Conclusions}}\label{sec:conclusions}

Ensuring robust extrinsic calibration is essential for any autonomous vehicle, as sensor drift arising from mechanical vibrations or changing operating conditions can otherwise compromise system performance. To address this, we introduce RLCNet, an end-to-end deep learning framework for online extrinsic calibration across LiDAR, RADAR, and camera sensors. RLCNet leverages convolutional neural networks (CNNs) to extract features from RGB images, LiDAR and RADAR projection images, and BEV maps. Correlation cost volumes are then constructed to capture spatial relationships indicative of sensor misalignment. A soft-mask feature-sharing mechanism strengthens inter-sensor connectivity prior to feature aggregation, which regresses transformation errors. Finally, a message-passing network (MPN) module refines predictions while enforcing loop-closure consistency.

RLCNet is trained with progressively decreasing miscalibration ranges through an iterative refinement strategy. The loss function integrates pose loss, point cloud distance loss, and loop-closure loss, with additional penalties on the MPN outputs to improve prediction consistency.

Unlike methods that regress absolute extrinsics, RLCNet predicts transformation errors, enabling continuous monitoring of sensor alignment. It supports real-time, frame-wise predictions and can trigger recalibration in response to sensor drift.

RLCNet achieves high calibration accuracy across all sensor pairs, with mean errors below $0.25^{\circ}$ in rotation and $1.5\,\text{cm}$ in translation. To ensure robustness in dynamic conditions, we further propose an online monitoring framework that exploits temporal coherence through weighted moving averaging and loop-closure-consistent recalibration. This system detects sensor drifts as small as $0.08^{\circ}$ and $1.6\,\text{cm}$ using only two consecutive frames.

\bibliographystyle{IEEEtran}
\bibliography{./bibliography/IEEEexample}

\end{document}